\documentclass[11pt,a4paper]{article}
\usepackage[file,caslon-inter]{paradigma}

\usepackage[english]{babel}
\usepackage{csquotes}
\usepackage[backend=biber,style=numeric,sorting=none]{biblatex}
\AtBeginBibliography{\raggedright}
\renewcommand*{\bibfont}{\pgcaptionfont}
\newfontfamily\pgpaperface{texgyretermes}[Extension=.otf,
  UprightFont=*-regular,BoldFont=*-bold,ItalicFont=*-italic,
  BoldItalicFont=*-bolditalic,Ligatures=TeX]
\setmathfontface\mathrm{lmroman10-regular.otf}[version=iclr]

\makeatletter
\ifpg@realfonts
  \newfontfamily\pgpapermono{CourierPrime}[Path=fonts/,Extension=.ttf,
    UprightFont=*-Regular,BoldFont=*-Bold,ItalicFont=*-Italic,
    BoldItalicFont=*-BoldItalic,Scale=0.88]
\else
  \newfontfamily\pgpapermono{texgyrecursor}[Extension=.otf,
    UprightFont=*-regular,BoldFont=*-bold,ItalicFont=*-italic,
    BoldItalicFont=*-bolditalic,Scale=0.88]
\fi
\newsavebox{\pg@synbox}
\newlength{\pg@synpad}
\newcommand{\synopsisrule}{\par\vspace{1.2ex}%
  {\color{pgtwilight!30}\rule{\linewidth}{0.4pt}}\par\vspace{0.6ex}\small}
\newcommand{\synopsislink}[2]{\par\noindent{\bfseries #1:}~#2}
\newcommand{\pgpaperhead}{\par
  \vspace*{0pt plus 1fill}%
  {\centering
    {\fontsize{17}{21}\selectfont\bfseries\pg@title: \pg@subtitle\par}%
    \vspace{2.2ex}%
    {\normalsize\pg@authors\par}%
    \vspace{0.8ex}%
    {\normalsize Paradigma\par}}%
  \vspace{26pt plus 1fill}}
\DeclareCaptionFont{pgquiet}{\fontsize{8}{9.5}\selectfont\color{pgink!80}}
\floatstyle{ruled}\restylefloat{algorithm}
\algrenewcommand{\algorithmicrequire}{\textbf{Input:}}
\newcommand{\pgiclrbody}{%
  \newgeometry{textwidth=5.5in,textheight=9in,centering,
    headheight=14pt,headsep=24pt,footskip=30pt}%
  \pagecolor{white}\color{black}%
  \colorlet{pgpaper}{white}\colorlet{pgink}{black}%
  {\pgpapermono\xdef\ttdefault{\f@family}}%
  \pgpaperface\edef\rmdefault{\f@family}%
  \mathversion{iclr}%
  \renewcommand{\normalsize}{\fontsize{10}{11}\selectfont}%
  \renewcommand{\small}{\fontsize{9}{10}\selectfont}%
  \renewcommand{\footnotesize}{\fontsize{9}{10}\selectfont}%
  \renewcommand{\large}{\fontsize{12}{14}\selectfont}%
  \renewcommand{\pgbody}{\normalfont\normalsize}%
  \renewcommand{\pgcaptionfont}{\normalfont\fontsize{8}{10}\selectfont}%
  \let\pgdisplayface\pgpaperface
  \let\pgserifface\pgpaperface
  \let\pglabelface\pgpaperface
  \renewcommand{\pglabelfont}{\normalfont\small}%
  \pgbody
  \setlength{\parindent}{0pt}\setlength{\parskip}{6pt}%
  \setlength{\abovedisplayskip}{7pt plus 2pt minus 5pt}%
  \setlength{\belowdisplayskip}{7pt plus 2pt minus 5pt}%
  \setlength{\abovedisplayshortskip}{0pt plus 3pt}%
  \setlength{\belowdisplayshortskip}{4pt plus 3pt minus 3pt}%
  \setlist{topsep=4pt plus 1pt minus 2pt,
    itemsep=2pt plus 1pt minus .5pt,parsep=2pt plus 1pt minus .5pt}%
  \titleformat{\section}{\normalfont\fontsize{17}{20}\selectfont\scshape
    \raggedright}{\thesection}{0.8em}{}%
  \titlespacing*{\section}{0pt}{0pt}{3ex plus .5ex minus .2ex}%
  \titleformat{\subsection}{\normalfont\fontsize{13}{16}\selectfont\scshape
    \raggedright}{\thesubsection}{0.8em}{}%
  \titlespacing*{\subsection}{0pt}{2.5ex plus .5ex minus .2ex}
    {1ex plus .2ex}%
  \pgsectionpagetrue
  \titleformat{\subsubsection}[hang]
    {\normalfont\normalsize\scshape\raggedright}{\thesubsubsection}{1em}{}%
  \titlespacing*{\subsubsection}{0pt}{1.5ex plus .5ex minus .2ex}
    {.5ex plus .2ex}%
  \titleformat{\paragraph}[runin]{\normalfont\normalsize\bfseries}{}{0pt}{}%
  \titlespacing*{\paragraph}{0pt}{1.5ex plus .5ex minus .2ex}{1em}%
  \renewenvironment{synopsis}
    {\par\begin{lrbox}{\pg@synbox}%
      \begin{minipage}{\dimexpr\linewidth-2\pg@synpad-1pt\relax}%
      {\centering\normalfont\large\scshape\color{pgtwilight}Abstract\par}%
      \vspace{1.4ex}\normalfont\normalsize}
    {\end{minipage}\end{lrbox}%
      \par\noindent\begin{tikzpicture}
        \node[draw=pgtwilight!55, fill=pgtwilight!5, line width=0.5pt,
          rounded corners=6pt, inner sep=\pg@synpad]{\usebox{\pg@synbox}};
      \end{tikzpicture}\par}%
  \captionsetup{font=small,labelfont=bf,labelformat=default,
    labelsep=colon,justification=justified,singlelinecheck=true}%
  \captionsetup[table]{position=top}%
  \renewcommand*{\bibfont}{\normalfont\normalsize}%
  \renewcommand{\topfraction}{.9}\renewcommand{\bottomfraction}{.8}%
  \renewcommand{\textfraction}{.07}\renewcommand{\floatpagefraction}{.8}%
  \setcounter{topnumber}{3}\setcounter{bottomnumber}{2}%
  \setcounter{totalnumber}{4}%
  \setlength{\@fptop}{0pt}\setlength{\@fpsep}{12pt plus 2fil}%
  \setlength{\@fpbot}{0pt plus 1fil}%
  \setlength{\headwidth}{\textwidth}%
  \fancypagestyle{paradigma}{\fancyhf{}%
    \renewcommand{\headrulewidth}{0pt}\renewcommand{\footrulewidth}{0pt}%
    \fancyhead[L]{\pglockup[3.6mm]}%
    \fancyhead[C]{\normalfont\small\pg@title}%
    \fancyhead[R]{\normalfont\small\pg@date}%
    \fancyfoot[C]{\normalfont\normalsize\thepage}}%
  \pagestyle{paradigma}%
}
\makeatother

\newif\ifpgsectionpage
\newif\ifpgsectionrunon
\AddToHook{cmd/section/before}{\ifpgsectionpage
  \ifpgsectionrunon\global\pgsectionrunonfalse
  \else\clearpage\suppressfloats[t]\fi\fi}
\AddToHook{cmd/appendix/after}{\global\pgsectionrunontrue}

\makeatletter
\edef\pg@contentpath{\detokenize{content}}%
\def\pg@sectioneof{\ifx\CurrentFilePath\pg@contentpath\clearpage\fi}
\AddToHook{file/before}{\ifx\CurrentFilePath\pg@contentpath
  \everyeof{\pg@sectioneof}\fi}
\AddToHook{file/after}{\ifx\CurrentFilePath\pg@contentpath\everyeof{}\fi}
\makeatother

\AddToHook{cmd/paradigmacover/after}{\pgiclrbody}

\pgtitle{Limite 1B - Violetto}
\pgsubtitle{Technical Report}
\hypersetup{pdftitle={Limite 1B - Violetto: Technical Report}}
\pgauthors{Mario Prignano, Gabriele Cirillo, Alessio Morosini, Luca Cerovaz,
  Alessandro Bartolocci\newline Emanuele Rodolà, Giulio Starace, Francesco Pappone}
\pgdate{October 2026}

\subjectline{Paradigma Research}{Technical report}
\filenote{}

\IfFileExists{fonts/CourierPrime-Regular.ttf}{%
  \newfontfamily\pgkeepfontA{fonts/CourierPrime-Regular.ttf}%
  \newfontfamily\pgkeepfontB{fonts/CourierPrime-Bold.ttf}%
  \newfontfamily\pgkeepfontC{fonts/CourierPrime-Italic.ttf}%
  \newfontfamily\pgkeepfontD{fonts/CourierPrime-BoldItalic.ttf}%
  \newfontfamily\pgkeepfontE{fonts/LibreCaslonDisplay-Regular.ttf}%
}{}

\begin{document}

\pgiclrbody
\IfFileExists{assets/paradigma-mark.png}{\setbox0=\hbox{\includegraphics[height=1pt]{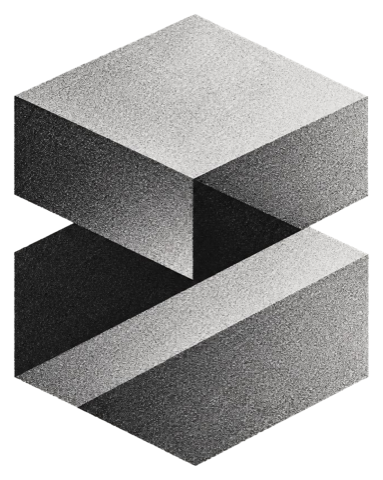}}}{}

\pgpaperhead
\begin{synopsis}
We present Limite 1B - Violetto, a one-billion-parameter model designed for high-throughput mathematical reasoning. Limite is trained from scratch on 237 billion tokens from a carefully curated, mathematics-centric corpus and supports context lengths of up to 131k tokens. Its architecture and training recipe were developed together to maximize reasoning capability while keeping training and inference costs low. Post-training combines knowledge distillation, targeted supervised fine-tuning, and RLVR on a curated curriculum. Limite achieves 94.01\% on AIME 2026 and 74.25\% on BeyondAIME \cite{maa_aime, bytedance_seed_2025_beyondaime}, placing it among models many times its size on our evaluation suite. The model is deliberately only lightly instruction-tuned: it is intended primarily as a single-turn mathematical solver and as a fast, parallel component of larger systems, rather than as a general-purpose assistant. We release the model, its base and value checkpoints, and an optimized serving implementation under the Apache 2.0 license.
\synopsisrule
\synopsislink{GitHub}{\url{https://github.com/paradigma-inc/limite-violetto}}
\synopsislink{Hugging Face}{\url{https://huggingface.co/paradigma-inc/models}}
\synopsislink{Blog}{\url{https://paradigma.inc/blog/limite-1b-violetto/}}
\synopsislink{Correspondence}{\href{mailto:research@paradigma.inc}{\nolinkurl{research@paradigma.inc}}}
\end{synopsis}

\vspace{26pt plus 1fill}
\begin{center}
\captionsetup{type=figure,font=pgquiet,width=325pt}%
\includegraphics{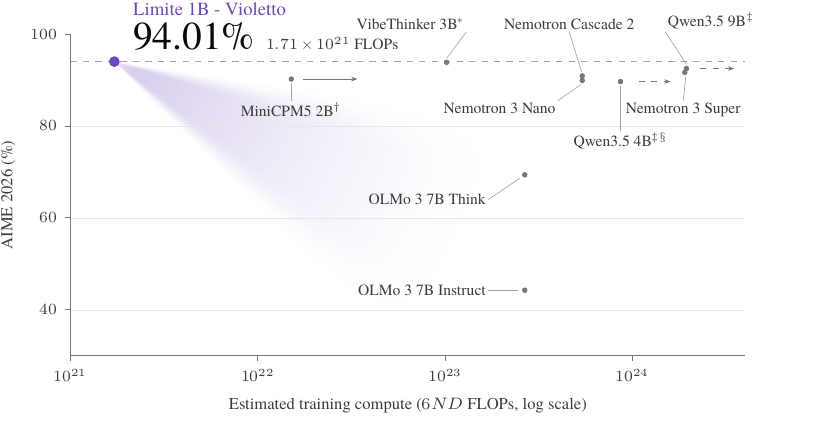}
\caption{AIME 2026 accuracy against estimated training compute. Limite 1B -
Violetto reaches 94.01\% at $1.71\times10^{21}$ FLOPs, one to three orders of
magnitude below the models that match it. Appendix~\ref{app:compute} explains
the compute estimates, the score sources and the marked points.}
\label{fig:compute}
\end{center}

\section{Introduction}\label{sec:introduction}
\subjectline{Introduction}{Claim and boundary}

Progress in mathematical reasoning has often come with increases in model size, training compute, and test-time sampling. While this has produced increasingly capable systems, it also makes repeated sampling, long-context inference, and large-scale deployment expensive. Many applications require a different operating point: a model that can solve difficult problems while remaining small enough to run at relatively high throughput and to generate many candidate solutions in parallel. This motivates a complementary question to conventional scaling: how much mathematical capability can we concentrate into a model with approximately one billion parameters?

We introduce \textbf{Limite 1B - Violetto}, a dense autoregressive model built around this objective. Limite is trained from scratch for mathematics rather than adapted from a general-purpose base model. With one billion parameters, it supports contexts of up to 131k tokens, and is designed to provide high reasoning capability per parameter. The model is only lightly instruction-tuned and is intended primarily as a single-turn mathematical solver, as opposed to a general conversational assistant.

Efficiency guides the design of the entire training pipeline. Limite uses a deep architecture with a mixture of local and global attention, allowing it to perform many sequential transformations while limiting the cost of long-context inference. The model is pre-trained on 237 billion tokens drawn from a larger mathematics-centric corpus containing problem-solution pairs, synthetic mathematical material, and carefully filtered STEM-adjacent web text. The resulting base model already exhibits substantial mathematical ability before post-training.

Post-training is organized as a sequence of complementary stages. We begin with knowledge distillation from millions of filtered mathematical solutions, then alternate targeted supervised fine-tuning with RL. Rather than repeatedly training on broad, static datasets, each stage focuses on areas where the current checkpoint remains weak. Later RL stages also use a learned value model to improve credit assignment over long reasoning traces.

We evaluate Limite across seven benchmarks covering recent mathematical competitions and research-level problems. It obtains 94.01\% on AIME 2026, 83.62\% on HMMT February 2026, and 74.25\% on BeyondAIME \cite{maa_aime, dekoninck2026benchmarksmatharenaevaluationplatform, bytedance_seed_2025_beyondaime}. A more comprehensive evaluation is available in Section~\ref{sec:evaluation}. These results are competitive with models ranging from several billion to tens of billions of parameters.

We envision Limite’s small size as possibly enabling a broader mode of use, where a controller may delegate mathematical questions to multiple Limite instances, inspect their outputs asynchronously, and combine promising lines of reasoning. This separates high-level coordination from high-frequency mathematical exploration and makes it possible to spend additional test-time compute through parallelism rather than through a single large model. The same specialization that makes Limite effective in this setting is also its principal limitation: outside mathematics, it may misinterpret a request or replace it with a superficially related mathematical task, an example of this is shown in Appendix \ref{app:qualitative-examples}. Also within mathematics, Limite works best at mathematical reasoning, while it may struggle with knowledge of the literature or formalization and structured theorem proving.

We show that strong mathematical reasoning can be achieved with a restricted parameter count and relatively few steps of training. By jointly designing the architecture, data, optimization, and post-training process around a narrow objective, Limite concentrates substantial reasoning capability into a compact model that can be sampled from and deployed efficiently. We release the final model, its base and value checkpoints, our evaluation suite, and the serving implementation all under the Apache 2.0 license. 

\section{Model Architecture}\label{sec:architecture}
\subjectline{Model Architecture}{Design record}

We designed Limite 1B to be as sample efficient as possible in its learning, without materially reducing training throughput. We evaluated architectural variants at shorter token horizons and selected a configuration that remained stable and efficient with our training settings.

Many architectural modifications are motivated primarily by converged model quality or inference efficiency, with less attention paid to wall-clock training time. We nevertheless still make some tweaks to reduce wall-clock overhead, inspired by modded nanogpt speedruns. Most additions are therefore applied only to selected layers or restricted portions of the residual stream, making them sparse in depth, width, or both. This keeps training throughput close to that of an otherwise standard modern Transformer. We detail our architecture in the following subsections and present a visual overview in Figure \ref{fig:arch}.

\plate{\makebox[\linewidth]{\includegraphics[width=\textwidth]{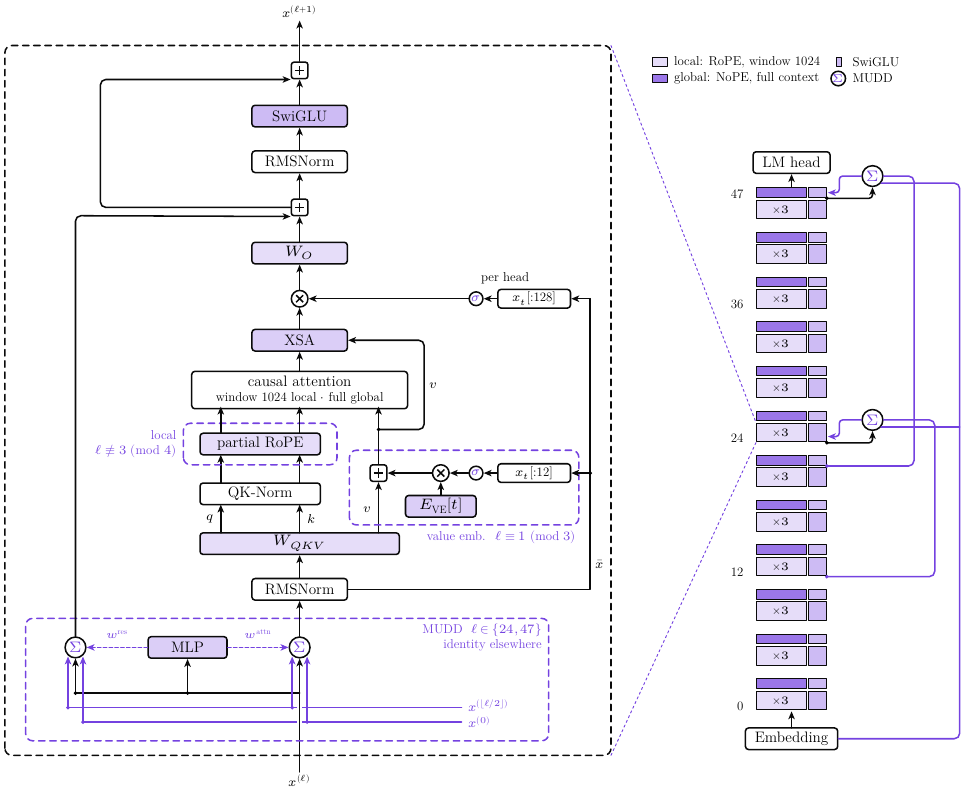}}}{%
  \label{fig:arch}%
  Architecture of Limite 1B - Violetto. Left: a Transformer block $\ell$, showing the
  simplified MUDD mixers at its input, gated value embeddings on the value
  branch, and Partial RoPE on local attention \cite{kazemnejad2023impactpositionalencodinglength, yang2025ropenopeagainnew}. Right: the 48-layer stack,
  arranged as 12 repetitions of three local-attention layers followed by one
  global-attention layer. Value embeddings are applied every third layer
  ($\ell \equiv 1 \pmod{3}$); they are omitted from the stack view for clarity.}

\subsection{Backbone and Block Structure}

Limite 1B is a 48-layer RMS-pre-norm Transformer \cite{vaswani2017attention, touvron2023llamaopenefficientfoundation, zhang2019rootmeansquarelayer} with a hidden dimension of $d=1280$ and tied embeddings \cite{press2017usingoutputembeddingimprove}. It uses a repeating pattern of three local-attention \cite{beltagy2020longformerlongdocumenttransformer, jiang2023mistral7b, child2019generatinglongsequencessparse} layers followed by one global-attention layer \cite{gemma2024gemma2, gemmateam2025gemma3technicalreport}. The local layers have a window size of 1024 and use Partial RoPE \cite{khan2026fractionalrotationpotentialinvestigating, su2023roformerenhancedtransformerrotary,black2022gptneox} with a base of 1024, whereas the global layers use NoPE \cite{yang2025ropenopeagainnew, kazemnejad2023impactpositionalencodinglength}. This hybrid pattern reduces KV-cache requirements while retaining regular access to the full context.

The model is considerably deeper than other Transformers of a similar size. This choice reflects our hypothesis that difficult mathematical reasoning benefits from a larger number of sequential operations, represented here through network depth~\cite{ye2025physicslm21,langlais2026itsalltraining}.

\paragraph{Logits soft-capping}
Finally, to make training more stable, the output logits are soft-capped with a shifted sigmoid
$$\tilde{z} = 23\,\sigma\big((z+5)/7.5\big)$$ where $z$ is produced by the tied
unembedding projection applied to the output of the last layer, a variant from the
modded-nanogpt speedrun~\cite{modded_nanogpt_2024} of the tanh
logit soft-capping of Gemma~2~\cite{gemma2024gemma2}.

\subsection{MLP}
The feed-forward layer is a standard bias-free SwiGLU \cite{shazeer2020gluvariantsimprovetransformer,elfwing2017sigmoidweightedlinearunitsneural} MLP without activation clamping. Given a residual state $x_t \in \mathbb{R}^{d}$ and its normalized representation
\[
\bar{x}_t = \operatorname{RMSNorm}(x_t),
\]
the MLP output is
\[
m_t
=
W_d
\left(
    \operatorname{SiLU }(W_g \bar{x}_t)
    \odot
    W_u \bar{x}_t 
\right),
\]
where
\[
W_g, W_u \in \mathbb{R}^{d_{\mathrm{ff}} \times d},
\qquad
W_d \in \mathbb{R}^{d \times d_{\mathrm{ff}}},
\qquad
d_{\mathrm{ff}} = 3328.
\]
The output is added to the residual stream using learned per-layer scalars:
\[
x_t
\leftarrow
\lambda_{\mathrm{mlp}} x_t
+
\lambda'_{\mathrm{mlp}} m_t.
\]

\subsection{Attention}

Each attention layer uses gated Grouped-Query Attention \cite{ainslie2023gqatraininggeneralizedmultiquery} with 10 query heads and 2 key-value (KV) heads. We augment the standard attention computation with gated value embeddings \cite{modded_nanogpt_2024}, Exclusive Self-Attention (XSA) \cite{zhai2026exclusiveselfattention} and gated attention \cite{qiu2025gatedattentionlargelanguage}.

Let $x_t \in \mathbb{R}^{d}$ denote the residual state at position $t$, and let
\[
\bar{x}_t = \operatorname{RMSNorm}(x_t)
\]
be the normalized input to the attention layer. For clarity, we describe the computation for a single head, with head dimension $d_h=128$. Queries, keys, and values are obtained as
\[
q_t = W_q \bar{x}_t,
\qquad
k_t = W_k \bar{x}_t,
\qquad
v_t = W_v \bar{x}_t.
\]
The queries and keys are then normalized as in \textcite{henry2020querykeynormalizationtransformers} and positionally encoded:
\[
\hat{q}_t
=
\mathrm{PE}_{\ell}\!\left(\operatorname{RMSNorm}(q_t)\right),
\qquad
\hat{k}_t
=
\mathrm{PE}_{\ell}\!\left(\operatorname{RMSNorm}(k_t)\right),
\]
where $\mathrm{PE}_{\ell}$ denotes Partial RoPE in local layers and the identity map in global layers.

\paragraph{Gated value embeddings}

Every third layer, specifically those satisfying $\ell \equiv 1 \pmod{3}$, augments the projected value with a learned embedding of the input token \cite{modded_nanogpt_2024}:
\[
\hat{v}_t
=
v_t
+
\alpha_t \odot \mathbf{V}_t.
\]
Here, $\mathbf{V}_t \in \mathbb{R}^{d_h}$ is retrieved from a learned lookup table
\[
\mathcal{V}
\in
\mathbb{R}^{N_{\mathrm{vocab}} \times d_h},
\]
using the token at position $t$. Its contribution is controlled by a per-token, per-head gate,
\[
\alpha_t
=
2\sigma\!\left(
    W_{\mathrm{gate}}\bar{x}_{t,1:12}
\right),
\]
which depends only on the first 12 channels of the normalized input \cite{modded_nanogpt_2024}. Layers without value embeddings use $\hat{v}_t=v_t$.

Standard causal attention is then applied to $\hat{q}_t$, $\hat{k}_t$, and $\hat{v}_t$, producing an output $z_t$ for each head.

\paragraph{Exclusive Self-Attention}

The attention output is subsequently processed by Exclusive Self-Attention \cite{zhai2026exclusiveselfattention}, which removes a learned fraction of its component aligned with the corresponding value vector, and later gated, effectively making our layer a form of gated attention \cite{qiu2025gatedattentionlargelanguage}, in practice:
\[
y_t
=
\beta_t
\left(
    z_t
    -
    \tanh(\gamma)
    \frac{ z_t^\top \hat{v}_t }
         {\lVert \hat{v}_t \rVert_2^2}
    \hat{v}_t
\right).
\]
The coefficient $\gamma$ is a learned scalar for each head. The output is also modulated by a per-token, per-head gate,
\[
\beta_t
=
2\sigma\!\left(
    W_{\beta}\bar{x}_{t,1:128}
\right),
\]
which reads only the first 128 channels of the normalized residual stream.

\subsection{Sparse Residual Mixing}

Modern Transformer architectures often modify the residual stream \cite{he2015deepresiduallearningimage} to increase communication between distant layers \cite{zhu2025hyperconnections,chen2026attnres,xie2026mhcmanifoldconstrainedhyperconnections}. Because a dense implementation would impose substantial training overhead, Limite uses a simplified MUDD-style residual mixer~\cite{xiao2025muddformer} at two locations in the network, effectively becoming an input-dependent form of a DenseFormer \cite{pagliardini2024denseformerenhancinginformationflow}.

At layers $\ell \in \{24,47\}$, the input to the Transformer block is replaced by a token-dependent linear combination of three residual states: the initial embedded representation, a state near the midpoint of the preceding stack, and the current layer input.

Let $x_t^{(\ell)}$ denote the residual state entering layer $\ell$, with
\[
x_t^{(0)}
=
\operatorname{RMSNorm}(\mathbf{E}_t),
\]
where $\mathbf{E}_t \in
\mathbb{R}^{d}$ is the token embedding at position $t$. At each mixing site, we define the set of residual taps
\[
\mathcal{T}_{\ell}
=
\left\{
    0,
    \left\lfloor \frac{\ell}{2} \right\rfloor,
    \ell
\right\}.
\]
The mixing weights are produced by a small two-layer MLP. Its first layer is shared across mixing sites:
\begin{gather*}
h_t
=
\operatorname{GELU}\!\left(
    W_1
    \operatorname{RMSNorm}(x_t^{(\ell)})
\right),
\qquad
W_1 \in \mathbb{R}^{32 \times d},
\\
w_t^{(\ell,r)}
=
W_2^{(\ell,r)} h_t
+
b^{(\ell,r)}
\in
\mathbb{R}^{3},
\qquad
r \in \{\mathrm{attn},\mathrm{res}\},
\\
\tilde{x}_t^{(\ell,r)}
=
\sum_{j \in \mathcal{T}_{\ell}}
w_{t,j}^{(\ell,r)} x_t^{(j)}.
\end{gather*}

Two mixtures are computed at each site. The first supplies the normalized input to the attention layer:
\[
\bar{x}_t^{(\ell,\mathrm{attn})}
=
\operatorname{RMSNorm}
\left(
    \tilde{x}_t^{(\ell,\mathrm{attn})}
\right).
\]
The second replaces the residual state onto which the attention output is added:
\[
x_t
\leftarrow
\lambda
\tilde{x}_t^{(\ell,\mathrm{res})}
+
\lambda' y_t.
\]

We initialize each second-layer matrix to zero and each bias to select the current residual state:
\[
W_2^{(\ell,r)} = 0,
\qquad
b^{(\ell,r)} = (0,0,1).
\]
Because $x_t^{(\ell)}$ is the final element of $\mathcal{T}_{\ell}$, the mixer is exactly the identity at initialization. The model therefore begins from the baseline architecture and learns to incorporate earlier residual states progressively throughout training.

This sparse formulation increases step time by only $2.3\%$ while retaining
most of the gain observed with the full MUDD formulation. At our scale, it
also performed comparably to the higher-overhead configurations of
Hyper-Connections~\cite{zhu2025hyperconnections} and Attention
Residuals~\cite{chen2026attnres} that we evaluated.

\section{Pretraining}

For the pre-training run of Limite, following the sample efficiency spirit that guided us for the whole work, we adopt very recent optimization algorithms and recipes, taking large inspiration from the Modded Nano GPT speedrun \cite{modded_nanogpt_2024}.
We pre-train Limite on 6 nodes of 8\times H100 \cite{nvidia2022h100}, for about 449k steps.

\subsection{Optimization}

We optimize the hidden
two-dimensional linear-operator weights with
NorMuon~\cite{li2025normuon}, with the exception of the simplified MUDD and gate linear layers.
NorMuon augments Muon's momentum orthogonalization~\cite{jordan2024muon} with
neuron-wise adaptive scaling based on an exponential moving average of
second-moment statistics. Algorithm~\ref{alg:normuon} gives the complete update
used in our final pretraining run.

We use Polar Express~\cite{amsel2026polarexpress} for the orthogonalization
step and retain Nesterov momentum \cite{1370862715914709505,       sutskever2013importance}. We additionally apply Cautious Weight
Decay~\cite{chen2026cautiousweightdecay} with a base coefficient of $1.2$ applied to the squared learning rate $\eta_t^2$, as in the modded-nanogpt optimizer; this
is equivalent to scaling the decay with the learning-rate schedule, as proposed
in AdamC~\cite{defazio2025gradientsrapidlyincreasenear}. This choice was intended to improve training stability and limit excessive weight growth, thereby reducing dynamic-range requirements that can complicate low-precision quantization~\cite{xiao2024smoothquantaccurateefficientposttraining}.

All remaining parameters are optimized with
AdamW~\cite{loshchilov2019decoupled}. These parameters are updated once every
two training steps, giving this optimizer group twice the effective batch size.
AdamW weight decay is disabled for all parameters except the tied token
embeddings and the value-embedding table.

We assign separate learning rates, momentum parameters, and weight-decay
multipliers to each parameter group. These settings were selected using
heuristics and shorter-horizon experimental runs and sweeps, and are reported in
Table~\ref{tab:optim}.

\begin{table}[H]
\centering
\small
\caption{Per-group optimizer hyperparameters, with Adam WD base $\lambda=0.005$ and a NorMuon base WD of 1.2, applied as $\eta^2\!\cdot\!\lambda\!\cdot\!\text{mult}$}
\label{tab:optim}
\begin{tabular}{llccc}
\toprule
\textbf{Parameters} & \textbf{Optimiser} & \textbf{Peak LR} & \textbf{Momentum / $(\beta_1,\beta_2)$} & \textbf{wd mult.} \\
\midrule
\texttt{$W_q, W_k, W_v, W_o$}   & NorMuon & $0.018$ & $(0.95,\,0.90)$ & $1$ \\
\texttt{$W_{gate}, W_{up}, W_{down}$}  & NorMuon & $0.018$ & $(0.95,\,0.90)$ & $1$ \\
\specialrule{\heavyrulewidth}{2pt}{2pt}
$E$ (tied)    & Adam & $0.0019$ & $(0.5,\,0.95)$  & $150$ \\
$\mathcal{V}$    & Adam & $0.1400$ & $(0.75,\,0.95)$ & $5$ \\
\text{scalars}            & Adam & $0.0400$ & $(0.9,\,0.99)$  & $0$ \\
\text{residual / post lambdas}      & Adam & $0.0016$ / $0.0080$ & $(0.9,\,0.95)$  & $0$ \\
\text{gates and XSA coefficients} & Adam & $0.0019$ & $(0.9,\,0.99)$  & $0$ \\
MUDD linears and biases & Adam & $0.0019$ & $(0.9,\,0.95)$ & $0$ \\
\bottomrule
\end{tabular}
\end{table}

\paragraph{Schedules}
All groups share one learning-rate schedule: a short linear warmup, a nearly
flat stable phase, and a linear decay to zero over the last 20\% of training.
Weight decay scales with $\eta_t^2$ and decays with it. NorMuon's momentum is
warmed up from 0.85 to 0.95; batch size and context length are fixed.

\paragraph{Pre-training run}
The final pre-training run used a batch size of 524k tokens. This relatively
small batch was part of our sample-efficiency-oriented training strategy. The
run remained stable for approximately 450k training steps, with no major
events such as large loss spikes.

Table~\ref{tab:architecture} summarizes the complete model configuration used
for pre-training and post-training.

\begin{table}[ht]
\centering
\small
\setlength{\tabcolsep}{5pt}
\newcommand{\bandrule}{\addlinespace[2pt]\arrayrulecolor{pgrule!45}\midrule[0.3pt]\arrayrulecolor{black}\addlinespace[2pt]}
\caption{Architecture of the Limite base model. Layer indices $\ell$ start
at 0. Parameter counts include the value-embedding table (38.8M); the input
and output embeddings are shared and counted once.}%
\label{tab:architecture}
\begin{tabularx}{\linewidth}{@{}Xl@{}}
\toprule
\multicolumn{2}{@{}l}{\bfseries Model} \\
\addlinespace[1pt]
Total parameters                       & 1.04B \\
Non-embedding parameters               & 0.80B \\
Layers                                 & 48 \\
Hidden size $d$                        & 1280 \\
Vocabulary size                        & 151{,}667 \\
Context length                         & 8192 in pretraining, 131k in post-training \\
\bandrule
\multicolumn{2}{@{}l}{\bfseries Attention} \\
\addlinespace[1pt]
Query heads / KV heads                 & 10 / 2 \\
Head dimension                         & 128 \\
Local layers                           & 36, sliding window 1024 \\
Global layers                          & 12 ($\ell \equiv 3 \bmod 4$), full context \\
Positional encoding (local / global)   & partial RoPE, 64 of 128 dims, base 1024 / NoPE \\
\bandrule
\multicolumn{2}{@{}l}{\bfseries Feed-forward} \\
\addlinespace[1pt]
Type                                   & SwiGLU, no bias \\
Intermediate size $d_{\mathrm{ff}}$    & 3328 ($2.6\,d$) \\
\bottomrule
\end{tabularx}
\end{table}

\subsection{Pre-training Data Curation}\label{sec:pretrain-data}

The composition of the pretraining corpus plays a central role in defining what
it means for the base model to be mathematics-specific. We wanted the
pretrained policy to acquire a strong mathematical prior before post-training,
while retaining sufficient linguistic variety to understand problems expressed
in different styles.

We therefore constructed a high-quality corpus that minimizes non-mathematical
content without sacrificing the breadth or volume of mathematical knowledge.
The corpus is divided into three source families:

\begin{itemize}
\item \textbf{Question and Answers (QA)}: problem--solution pairs, both
  synthetic and human-written, with and without explicit reasoning traces.
\item \textbf{Synthetic Knowledge (SK)}: synthetic mathematical documents,
  primarily written in the style of textbooks, lecture notes, and
  teacher--student conversations about specific mathematical topics.
\item \textbf{Filtered Web (FW)}: quality-filtered web pages, approximately
  40\% of which consists of synthetic rewrites of material from the same
  crawl.
\end{itemize}

The final mixture contains 46\% QA, 19\% SK, and 35\% FW.

We assessed data quality along several dimensions. These included the estimated
fraction of documents containing clear logical or mathematical errors and, for
the QA component, the correctness of the reported reference answers. Because
exhaustive verification was impractical, we used LLM-based checks on
subsamples from each source.
In practice for each source we draw uniform random subsamples of documents, and ask an LLM to assign a verdict, together with a written justification; for QA documents the reference
answer was independently re-derived and compared. Per-source error rates and
confidence intervals were then used to decide inclusion.

QA and SK provide most of the corpus's explicit mathematical knowledge and
reasoning content. FW serves a complementary role by increasing linguistic and
stylistic diversity. Its inclusion was carefully filtered to minimize logical
errors and common web-data artifacts, including formatting markup and
non-English text.

We intentionally included a substantial proportion of explicit reasoning
traces. Our aim was to provide a stronger initialization for the subsequent
post-training pipeline, motivated by growing controlled evidence that exposing
models to diverse reasoning patterns earlier in training can improve both base-model
reasoning capabilities and their ability to benefit from subsequent post-training
stages~\cite{akter2026frontloading,wang2025thinkingaugmentedpretraining,tan2026selfimprovingpretrainingusingposttrained,shen2026understandingreasoningpretrainingposttraining,langlais2026itsalltraining}.

We applied lexical decontamination followed by a lighter semantic
decontamination stage. The semantic filtering threshold was tuned separately
for each source family, and both stages were applied against every benchmark
reported in this work.

The majority of the corpus has a general knowledge cutoff around the final months of 2025.
Most non-synthetic web documents predate 2025, although a small fraction
extends into early 2026.

The final polished mixture contains approximately 346 billion tokens. The final
pretraining run consumed a 236 billion tokens subset of this mixture, producing
the checkpoint that we refer to as \emph{Base}. This checkpoint was then used
as the starting point for the post-training pipeline.

\subsection{Model Souping}

After pretraining, we experimented with model
souping~\cite{wortsman2022modelsoups}, in which checkpoints obtained from
related training runs are merged by averaging their parameters. Our objective
was to improve upon the base model using a comparatively small amount of
high-quality data, without collapsing the model's output diversity.

We constructed a mixture of approximately 6 billion tokens from the
highest-quality sources in the pretraining corpus together with selected
post-training data. A large proportion of these documents contain explicit
reasoning traces. Starting from the same base checkpoint, we trained three
independent runs in parallel, varying the data-shuffling seed and learning-rate
settings between runs. We then formed the final soup by uniformly averaging the
three resulting checkpoints \cite{olmo20252olmo2furious}.

As shown in Table~\ref{tab:soup-evals}, the souped model
outperforms each individual run and the final pretraining checkpoint on MATH-500 evaluation \cite{hendrycks2021measuringmathematicalproblemsolving, lightman2023letsverifystepstep}, and also
improves the model's reasoning capabilities.

We release this final soup together with the \emph{Base} checkpoint used to initialize
the post-training pipeline.

\begin{table}[h]
\centering
\pgcaptionfont
\newcommand\dlt[1]{\,\textcolor{pgtwilight}{#1}}
\setlength{\tabcolsep}{5pt}
\caption{MATH-500 greedy generation accuracy (\%) of the base checkpoint, the
three single runs and their uniform soup, with and without thinking. With
thinking, the prompt ends with an opening \texttt{<think>} token. Coloured
numbers give the change from thinking relative to the same model without it.
Bold marks the best value within each prompt setting.}%
\label{tab:soup-evals}
\begin{tabular}{@{}lrrr@{}}
\toprule
& \multicolumn{2}{c}{Prompt} & \\
\cmidrule(lr){2-3}
Model & Base & Think & $\Delta$ \\
\midrule
Limite Base V5       & 61.8 & 59.2 & \dlt{$-$2.6} \\
Single run A         & 67.4 & 73.0 & \dlt{+5.6} \\
Single run B         & 66.2 & 74.6 & \dlt{+8.4} \\
Single run C         & 64.0 & 71.6 & \dlt{+7.6} \\
\midrule
Uniform soup (A+B+C) & \textbf{68.4} & \textbf{80.2} & \dlt{+11.8} \\
\bottomrule
\end{tabular}
\end{table}

\subsection{Base-Model Evaluation}

Table~\ref{tab:pretrain-base-results} reports the mathematical-reasoning
results available for the base checkpoints before post-training. 

With thinking, the soup improves on the base checkpoint on MATH-500 and every
AIME year. GSM8K behaves differently: thinking lowers both models to about
46\%, and without thinking the soup trails the base by 17 points. We suspect
our MATH-style few-shot exemplars, which suit GSM8K poorly, and that the soup
is more sensitive to this format mismatch than the base checkpoint.

\begin{table}[ht]
\centering
\small
\setlength{\tabcolsep}{4pt}
\newcommand{\hd}[2]{\begin{tabular}[b]{@{}c@{}}#1\\#2\end{tabular}}
\caption{Accuracy (\%) of the base checkpoint and of the final soup, without and
with thinking. Few-shot prompts use MATH-style exemplars, which may not suit
GSM8K: 4-shot for MATH-500 and AIME, 8-shot for GSM8K; thinking appends an
opening \texttt{<think>} token. Sampling at $T=0.6$, top-$p=0.95$, seed 0,
8{,}192 generated tokens within a 12{,}288-token context. Responses per
problem: 4 for MATH-500, 1 for GSM8K, 32 for AIME; lenient scoring averaged
over all responses.}%
\label{tab:pretrain-base-results}
\begin{tabular}{@{}l*{5}{c}@{}}
\toprule
Model & \hd{MATH}{500} & GSM8K & \hd{AIME}{2024} & \hd{AIME}{2025} & \hd{AIME}{2026} \\
\midrule
Limite Base                  & 60.70 & \textbf{80.52} & 3.33  & 3.75  & 3.33 \\
Limite Base (thinking)       & 73.40 & 46.10 & 15.52 & 23.85 & 16.88 \\
Limite Base Soup             & 59.25 & 63.46 & 7.40  & 8.54  & 8.23 \\
Limite Base Soup (thinking)  & \textbf{81.50} & 46.32 & \textbf{33.33} & \textbf{32.92} & \textbf{34.06} \\
\bottomrule
\end{tabular}
\end{table}

\section{Post-Training}\label{sec:posttraining}
\subjectline{Post-Training}{Method record}

\plate{\makebox[\linewidth]{\includegraphics[width=\textwidth]{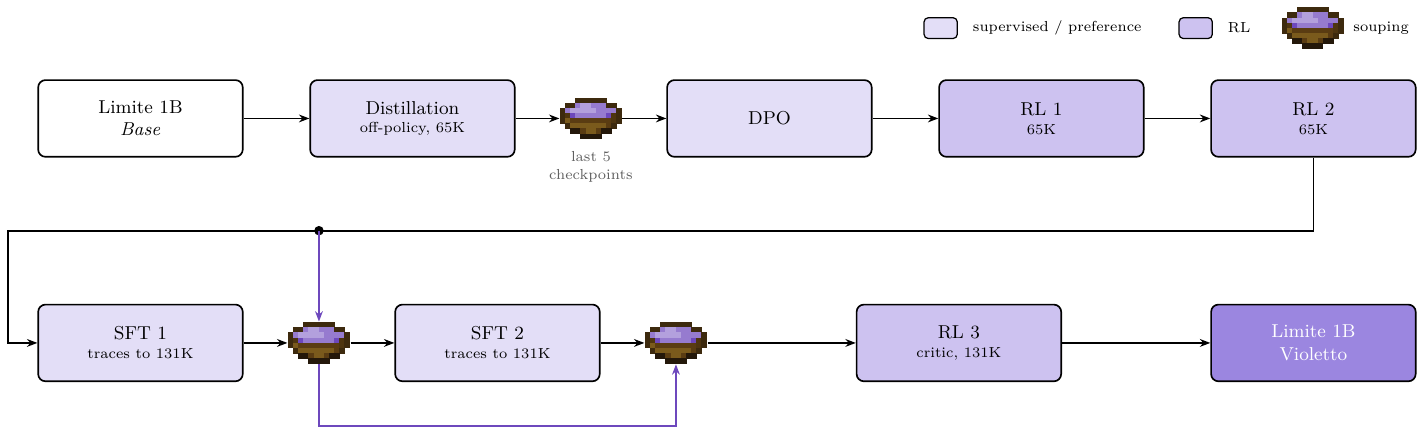}}}{%
  \label{fig:posttrain-pipeline}%
  The post-training pipeline, from Limite 1B \textit{Base} to Limite 1B - Violetto. We
  first distill the base model off-policy at a 65K context, average the last
  five distillation checkpoints, and apply DPO. Two RL runs follow, also at a
  65K context. A first short SFT stage then trains on longer traces of up to
  131K tokens, and its final checkpoint is souped with the checkpoint from the
  second RL run. A second short SFT stage, again on traces of up to 131K
  tokens, starts from this soup, and its final checkpoint is souped with the
  first soup. The last RL run starts from the second soup, uses a critic, and
  trains at a 131K context. Each bowl marks a souping step, a weighted average
  of checkpoints.}

While we find that the base model performs comparatively well with respect to similar base models, it nevertheless has never been exposed to long rollouts nor to its \textit{own} rollouts during training. We deem these as instrumental for downstream performance, and therefore orient our post-training on these directions.

Our post-training pipeline follows the same sample-efficiency goal as pre-training. It begins with a comparatively long knowledge-distillation~\cite{hinton2015distillation} (KD) stage followed by a short direct preference optimization~\cite{rafailov2024directpreferenceoptimizationlanguage} (DPO) run, then interleaves three reinforcement-learning runs with short, highly curated supervised fine-tuning (SFT) \cite{ouyang2022traininglanguagemodelsfollow, wei2022finetunedlanguagemodelszeroshot} stages and model souping. We view the SFT stages as introducing targeted capabilities, and the RL runs as sharpening them with high learning rates and relatively few expensive updates. Our RL procedure draws from DPPO~\cite{dppo}, DAPO~\cite{dapo} and VAPO~\cite{vapo}-related variants.

\subsection{Post-Training Pipeline}

Post-training has become a fundamental step of frontier model training, and apart from data, the algorithms used 
play an essential role in the downstream capabilities of the models. Research on this field recently accelerated and one can now choose from a plethora 
of solutions when assembling the pipeline.
We choose to initially distill from a teacher model and then combine highly 
curated short SFTs and souping with custom RL algorithms to push the model capability as much as we can, 
exploiting high learning rates despite being off-policy.

\paragraph{Knowledge distillation}
The first stage is teacher-forced, logit-level distillation on the corpus described in Section~\ref{par:distill}. At each position, the training target is the teacher's top 32 log-probabilities for its generated solution. We conclude by uniformly averaging the final five checkpoints~\cite{swa}.

\paragraph{Direct preference optimization}
After distillation, we apply one epoch of offline DPO~\cite{rafailov2024directpreferenceoptimizationlanguage}
with mean (length) normalization of the implicit reward, following
LN-DPO~\cite{ahrabian2025practicalanalysishumanalignment}. Both the trained
policy and the frozen reference policy are initialized from the distilled
soup. Averaging the per-token log-ratios, rather than summing them, keeps the
preference signal from favouring shorter or longer responses
(Appendix~\ref{app:dpo}). We train for 140 updates of 128 preference pairs
each, with $\beta=5$, using AdamW~\cite{loshchilov2019decoupled} with a peak
learning rate of $3\times10^{-6}$ and linear warmup and decay. No sequence is
truncated: every pair fits within the 65,536-token context.

\paragraph{Supervised fine-tuning and checkpoint averaging}
For SFT, we adapt the pretraining process to longer contexts while retaining optimizer and hyperparameters configurations, with the exception of the peak learning rate which we set to 15\% of the peak used in pre-training. Targeted mixtures and checkpoint averaging make these short stages effective. As outlined below, we interleave multiple RL and SFT phases. During the two post-RL SFT stages, we form one model soup between the stages and another before the final critic-assisted RL run. Each is an average of the current and preceding checkpoints~\cite{wortsman2022modelsoups}. These stages also extend the maximum context length from 65k to 131k tokens.

\paragraph{Interleaving SFT and RL}\label{par:sft-rl}

Continued RL eventually became unstable in our runs since we adopted a high learning rate, high off-policy regime. Since simply lowering the learning rate would worsen the sample efficiency of the already expensive updates, we opted for short SFT stages to stabilize trajectories while preserving capabilities acquired through RL. SFT introduces targeted information and raises pass@$k$ \cite{malladi2026tailsftfilteredfinetuningimproves, dang2025weightensemblingimprovesreasoning}, but can also move the model away from its on-policy distribution and weaken coherent sampling from that distribution. We found that averaging the pre- and post-SFT checkpoints balances these new capabilities against the sampling behaviour learned through RL; Following runs then continued from the resulting soup.

\subsection{Post-Training Data Curation}\label{sec:data}
To post-train a compact 1B-parameter model with substantially fewer tokens than models with similar math capabilities, we prioritize data quality across three sources: distillation targets, problems for RL, and problem--solution pairs for the short SFT stages.

\paragraph{Distillation corpus}\label{par:distill}

The knowledge-distillation corpus contains 5.5 million solutions generated by our main teacher, VibeThinker-3B \cite{xu2026vibethinker3bexploringfrontierverifiable}, for publicly sourced mathematical problems. These solutions comprise approximately 26 billion target tokens. At every position, the student is trained against the top 32 log-probabilities from the teacher distribution.

Public datasets may contain incorrect gold answers, damaged examples, or invalid solutions. We therefore apply three filters:
\begin{itemize}
    \item We retain a trace only if it reaches the problem's gold answer.
    \item We discard examples that are not mathematical problems or that arrive malformed.
    \item We discard traces in which the teacher abandons the derivation and guesses, identified through phrases such as ``given the time constraints'' or ``I'll just go with.''
\end{itemize}

The first filter removes incorrect traces and most examples with incorrect gold answers, but not cases in which the teacher happens to match an incorrect gold answer. We pass rejected traces to a judge model, which receives the problem and trace without the gold answer and determines whether the problem is well posed and the solution correct. This recovers 18\% of the corpus tokens that the first filter would otherwise reject.

Solutions contain fewer than 5k tokens on average and are capped at 65,536 tokens. Approximately one third of all tokens lie in a long-tail subset whose median solution length is around 15k tokens; the distillation schedule ends on this long-tail portion.

\paragraph{Targeted data selection}
Distillation provides an initially broad exposure to general mathematical problem solving; the subsequent bottleneck then remains performance on specific topics. We identify these topics through failure analysis and area-level accuracy, and direct later data generation toward the identified weaknesses. The short SFT stages consequently use an order of magnitude fewer tokens than distillation, concentrating on capabilities the current checkpoint does not yet exhibit reliably.

\paragraph{Preference pairs}
For the preference-optimization stage we use the mathematics subset of
Dolci-Think-DPO-32B~\cite{olmo2026olmo3}, whose pairs follow the delta-learning
construction~\cite{geng2025deltalearninghypothesispreference}: the chosen response is generated by
Qwen3-32B and the rejected response by Qwen3-0.6B, both in reasoning mode. We decontaminate following the same procedure followed in pre-training, described in Section~\ref{sec:pretrain-data}. We apply this same decontamination procedure to all of our post training work.

\paragraph{RL problem generation}\label{par:rl-problem-generation}
Rather than relying exclusively on public datasets, we build a dedicated generation pipeline that controls problem difficulty and schedules themes and mathematical areas across runs. Because an incorrect gold answer rewards mathematical errors and penalizes correct reasoning, every generated problem must have a programmatically computed answer and an empirical difficulty estimate from the checkpoint being trained. We use two complementary generators: an LLM-based generator and a CPU-based procedural generator.

The LLM-based generator produces a problem statement, a solving program, and a claimed answer. A validation gate re-executes the program in a separate environment and retains the problem only if its output matches the claimed answer. The generation request selects a mathematical area from a fixed list of subcategories, supplies a small set of examples, and asks for an unambiguous statement and a program implementing its conditions. Since the examples have the strongest influence on difficulty and register, the initial waves use a small, carefully curated set; later waves add generated examples whose measured difficulty falls in the useful range. Areas are sampled non-uniformly, concentrating generation on weaker areas. We outline the LLM-based generation process in Figure~\ref{fig:llm-generator}.

\begin{center}
\captionsetup{type=figure}
\includegraphics[width=\textwidth]{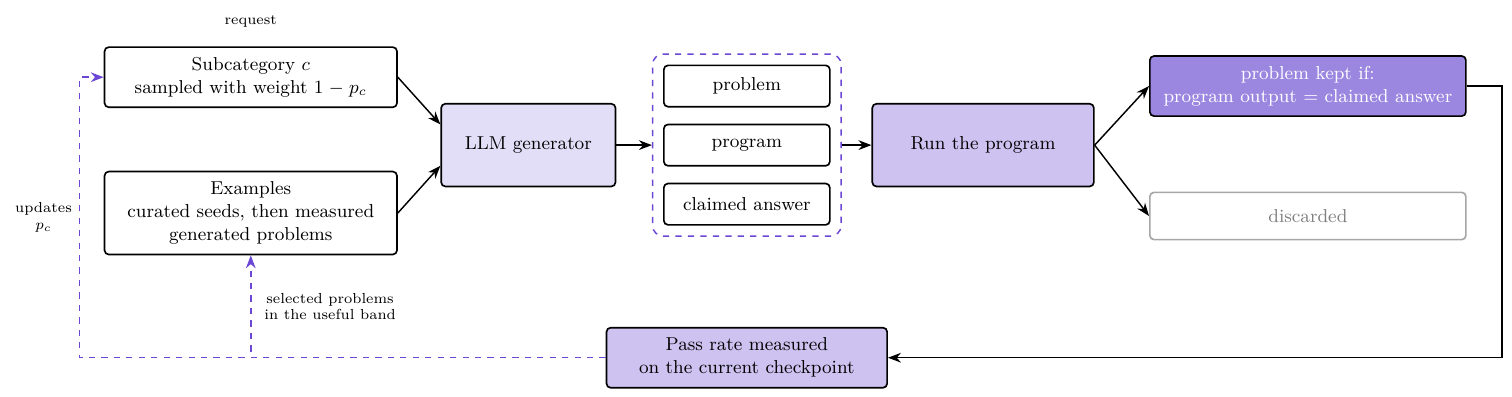}
\caption{Graphical representation of the LLM-based generator.}
\label{fig:llm-generator}
\end{center}

The complementary procedural generator is fully CPU-based and inspired by~\cite{saxton2019analysingmathematicalreasoningabilities}. This is suitable for RL, where thematic variety matters less than for SFT because GRPO~\cite{grpo} baseline subtraction removes signals shared across a rollout group. The generator combines high-quality program templates with keyword synonyms and alternative phrasings, preserving programmatic guarantees while adding linguistic variation. Each wave is strictly filtered and deduplicated; sampling hyperparameters are tuned per template, and a tuned sampler discourages trivial strategies such as direct enumeration in combinatorics problems. We outline the procedural generation process in Figure~\ref{fig:procedural-generator}.

\begin{center}
\captionsetup{type=figure}
\includegraphics[width=\textwidth]{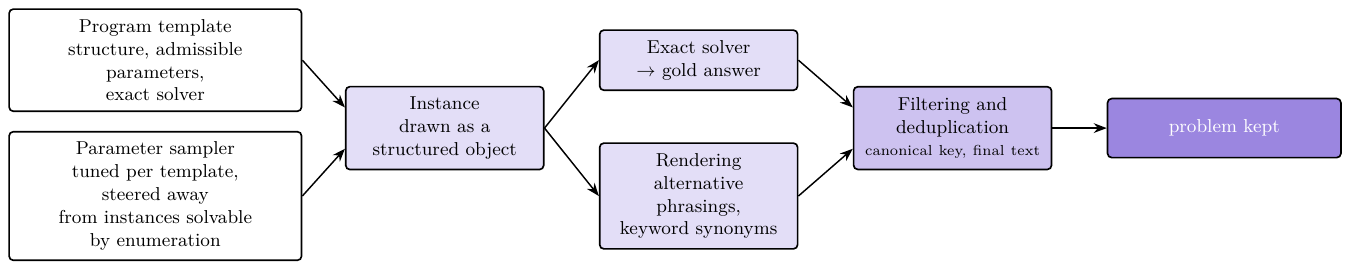}
\caption{Diagram explaining the procedural generator.}
\label{fig:procedural-generator}
\end{center}

The two generators offer complementary trade-offs. Once implemented, a procedural family can sample, solve, and render an instance in milliseconds on CPU, making volume inexpensive and limiting each wave primarily by the number of available families. By contrast, LLM-based generation may require tens of thousands of GPU-generated tokens per problem, but it can vary the underlying mechanism rather than only details and form. We therefore use procedural generation for volume and controllable difficulty, and LLM-based generation for structural variety.

After applying the same error analysis done on the public datasets, the generated sets yield a substantially lower rate of incorrect gold answers, approaching zero in our measurements.

\paragraph{Difficulty calibration}
Group-relative RL is informative only when the policy solves a problem intermittently: groups in which every rollout succeeds or every rollout fails have zero relative advantage. We therefore estimate pass rates using groups of eight rollouts from a fixed checkpoint and retain an intermediate difficulty band, favouring problems near a 50\% pass rate. For each wave, we analyze a representative sample large enough to estimate the difficulty distribution across families and subcategories; the result shifts subsequent sampling toward areas in which the current policy remains weaker \cite{baroian2026promptreplayspeedinggrpo, zhu2026rolloutpassratecontrolsteering, faro2026frontierlearningtrainingllm}.

\paragraph{Short-SFT mixtures}
By the end of the second RL run, reward had stopped improving, and failure analysis showed that the model remained weak in specific mathematical areas. Because we believe RL in our setting tends to only reinforce behaviors already represented by the policy \cite{yue2025doesreinforcementlearningreally, ICLR2026_517f9b9c}, we introduced two short SFT stages with different mixtures of the following two categories of sources:

\begin{itemize}
    \item Synthetic problems targeting the areas in which the current policy is weakest. The solutions are generated by Qwen3.8-27B and Qwen3.8-Flash-Next \cite{qwen38, qwen3.8flashnext} with a context length of 131k tokens.
    \item The remainder combines solutions generated by the current policy for our synthetic problems, which limit drift from the RL checkpoint, with a small proof subset. The proof subset is the union of proofs from public datasets and synthetic proofs for publicly available mathematical data, with the latter verified by an LLM judge. We include these proofs to introduce reasoning capabilities not yet present in the current policy.
\end{itemize}

\subsection{Reinforcement-Learning Method}
Reinforcement learning is central to reasoning-model post-training but expensive because rollouts are generated autoregressively. For Limite, the 48-layer architecture also gives a relatively high per-token latency despite its small parameter count. We therefore optimize for wall-clock efficiency through the following training choices:

\begin{itemize}
    \item \textbf{High learning rates} We use learning rates from $1\times10^{-5}$ to $2\times10^{-5}$, compared with the $1\times10^{-6}$ scale commonly used in comparable RL runs.
    \item \textbf{Many off-policy updates} We perform between 7 and 19 updates per behaviour policy, using the DPPO mask to control policy drift.
    \item \textbf{Few actor updates} We perform 461 RL updates across the three runs, rather than the thousands of updates often used in larger-scale training.
    \item \textbf{Interleaved SFT and RL} Short SFT stages introduce targeted capabilities and allow RL to resume from a different checkpoint.
    \item \textbf{Critic-assisted credit assignment} After the policy has learned the easiest improvements available through group-relative training, we introduce a critic for the final RL run.
\end{itemize}

The first two runs use the group-mean advantage estimate associated with GRPO~\cite{grpo}; the third uses a learned value model to assign token-dependent advantages. The credit-assignment and critic paragraphs below motivate this transition, followed by the complete training sequence.

\paragraph{Objective, reward, and dynamic sampling}

Let $o_i$ be a response generated by policy $\pi_\theta$ for query $q_i$. We use a sparse terminal reward at token $t$:

\begin{equation}
\label{eq:posttraining-reward}
R_{i,t} =
\begin{cases}
    1, & \text{if } t=\lvert o_i\rvert \text{ and } o_i \text{ has a correct final answer}, \\
    0, & \text{otherwise}.
\end{cases}
\end{equation}

For each problem, we sample $G$ rollouts and apply dynamic sampling, introduced by DAPO~\cite{dapo}, so that groups with pass rate $0/G$ or $G/G$ are discarded because all their group-relative advantages are zero. The resulting batches contain only groups that provide a learning signal.

To maximize the number of off-policy updates while retaining training stability, we combine the DPPO masking mechanism with CISPO~\cite{cispo}-style clipping of the importance sampling weight. For rollout $j$ associated with query $i$, define
\[
r_{i,j,t}
=
\frac{\pi_\theta(o_{i,j,t}\mid q_i,o_{i,j,<t})}
     {\mu(o_{i,j,t}\mid q_i,o_{i,j,<t})},
\]
where $\mu$ is the behaviour policy. Let $\hat A_{i,j,t}$ be the advantage and $M_{i,j,t}$ the DPPO mask. The token-normalized policy loss is
\begin{equation}
\label{eq:posttraining-policy-loss}
\mathcal{L}_\mathrm{policy}(\theta)
=
\frac{
\displaystyle\sum_{i=1}^{B}\sum_{j=1}^{G}
\sum_{t=1}^{\lvert o_{i,j}\rvert}
M_{i,j,t}\hat A_{i,j,t}\widetilde r_{i,j,t}
}{
\displaystyle\sum_{i=1}^{B}\sum_{j=1}^{G}
\lvert o_{i,j}\rvert
}.
\end{equation}
Here, $M_{i,j,t}=1$ includes the token and $M_{i,j,t}=0$ masks it. The quantity $\widetilde r_{i,j,t}$ has the same forward value as $r_{i,j,t}$ but a clipped gradient. We normalize by tokens rather than samples~\cite{dapo}; both choices introduce length-dependent biases, and for these runs we prefer the token-level bias.

\paragraph{Rollout batching}
\label{par:roll-batch}
We considered two ways to overlap rollout generation and training:
\begin{enumerate}
    \item As soon as every rollout in a group finishes, place the group in a buffer if it is informative and its behaviour policy is sufficiently recent. When the buffer is full, update the target policy while generation continues.
    \item Generate a cohort of $B$ complete groups, discard non-informative groups, split the remainder into minibatches, and then update on those minibatches.
\end{enumerate}

The first approach is faster, but it preferentially admits groups that finish early. This creates two potential biases:
\begin{itemize}
    \item \textbf{Length bias} Training more often on short rollouts may encourage the model to shorten its reasoning.
    \item \textbf{Problem-difficulty bias} Problems solved quickly are often easier, so early admission may overrepresent easier problems.
\end{itemize}

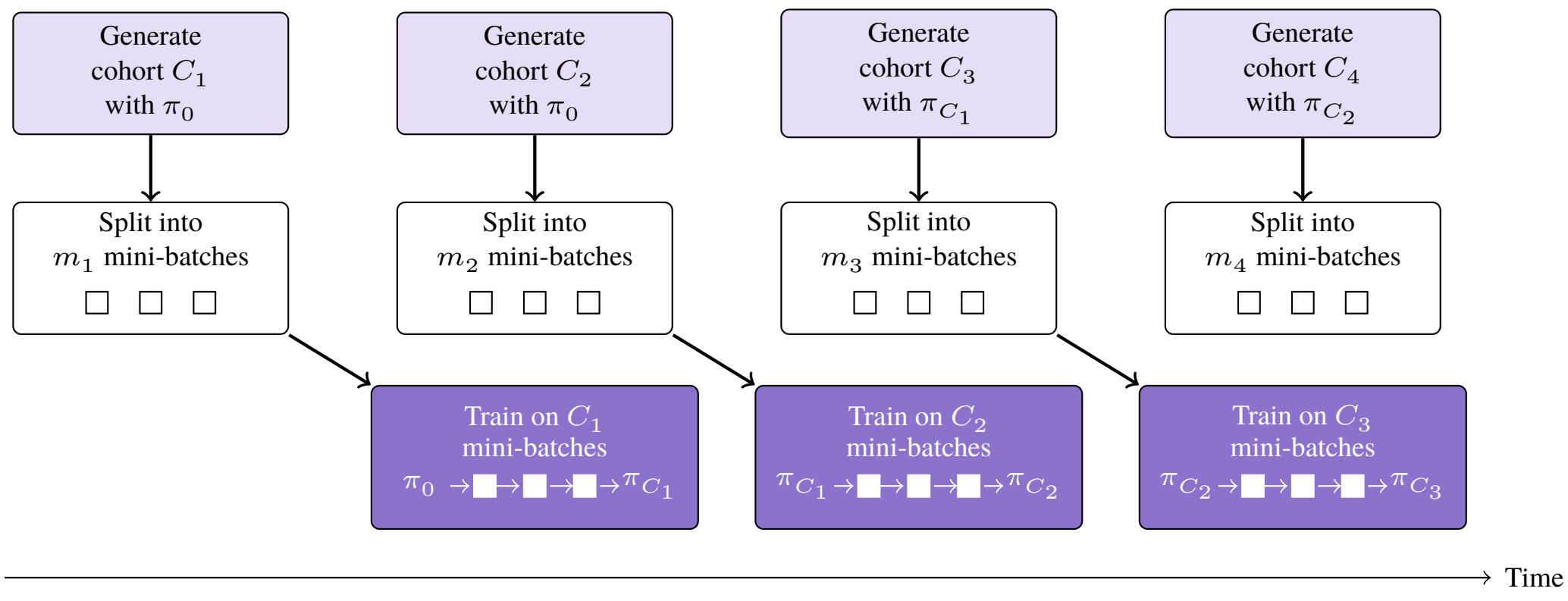
\begin{figure}[htbp]
\noindent\makebox[\linewidth][l]{%
\definecolor{genviolet}{HTML}{E8DDF7}
\definecolor{trainviolet}{HTML}{B99BE5}
\begin{tikzpicture}[x=3.35cm, y=1cm, font=\fontsize{7pt}{8.5pt}\selectfont]

\begin{scope}[xshift=-0.4cm]

  \foreach \x/\k/\pol in {
    0/1/0,
    1/2/0,
    2/3/C_1,
    3/4/C_2
  } {
    \node[draw, rounded corners=2pt, fill=genviolet,
          minimum width=2.4cm, minimum height=1.0cm,
          align=center]
          (gen\k) at (\x+0.5,3.85)
          {Generate\\cohort $C_{\k}$\\with $\pi_{\pol}$};

    \node[draw=black, rounded corners=2pt, fill=white,
      minimum width=2.4cm, minimum height=1.15cm]
      (split\k) at (\x+0.5,2.15) {};

    \node[text=black, font=\fontsize{7pt}{8.5pt}\selectfont, align=center]
          at (\x+0.5,2.38)
          {Split into\\$m_{\k}$ mini-batches};
    
    \foreach \d in {-0.14,0,0.14} {
      \node[draw=black, fill=white,
            minimum size=0.19cm, inner sep=0pt]
            at ({\x+0.5+\d},1.85) {};
    }

    \draw[->, thick] (gen\k.south) -- (split\k.north);
  }

  \definecolor{trainviolet}{HTML}{8B73CC}

\foreach \x/\k/\old/\new in {
  1/1/0/C_1,
  2/2/C_1/C_2,
  3/3/C_2/C_3
} {
  \node[draw, rounded corners=2pt, fill=trainviolet,
        minimum width=2.85cm, minimum height=1.25cm]
        (train\k) at (\x+0.5,0.5) {};

  \node[text=white] at (\x+0.5,0.84) {Train on $C_{\k}$};
  \node[text=white] at (\x+0.5,0.59) {mini-batches};
  \node[text=white] at (\x+0.20,0.25) {$\pi_{\old}$};

  \foreach \d in {0.37,0.50,0.63} {
    \node[draw=white, fill=white,
          minimum size=0.19cm, inner sep=0pt]
          at ({\x+\d},0.25) {};
  }

  \node[text=white] at (\x+0.80,0.25) {$\pi_{\new}$};

  \draw[->, white] (\x+0.28,0.25) -- (\x+0.33,0.25);
  \draw[->, white] (\x+0.40,0.25) -- (\x+0.46,0.25);
  \draw[->, white] (\x+0.54,0.25) -- (\x+0.60,0.25);
  \draw[->, white] (\x+0.67,0.25) -- (\x+0.72,0.25);

  \draw[->, thick] (split\k.south east) -- (train\k.north west);
}

\end{scope}

  \draw[->] (0,-0.55) -- (3.87,-0.55) node[right] {Time};

\end{tikzpicture}%
}
\caption{Double buffering across cohorts. $C_k$ denotes the $k$-th
cohort and $m_k$ the number of mini-batches obtained from it. The
white squares represent mini-batches schematically. The trainer performs one update per
mini-batch while the next cohort is generated. $\pi_0$ is the initial
policy, and $\pi_{C_{k+1}}$ is the policy $\pi_{C_k}$ after completing $m_{k+1}$ updates
on $C_{k+1}$ mini-batches.}
\label{fig:double-buffering}
\end{figure}

Both approaches overlap rollout generation and training, the only difference is how rollouts enter in a batch. The first one is similar to PRIME-RL's~\cite{prime_rl} asynchronous buffering, while the second one is closer to DAPO's~\cite{dapo} implementation.  We implement the second approach using double buffering, as illustrated in Figure~\ref{fig:double-buffering}, to obtain more representative cohort-level updates. Our double buffering setup resembles those of Asynchronous RLHF~\parencite{async_rlhf} and Cleanba~\parencite{cleanba}.

As shown in Figure~\ref{fig:double-buffering}, unless from the initial policy $\pi_0$, each policy trains on off-policy rollouts. The lag between the target policy $\pi_{C_{k+2}}$ and the behaviour policy $\pi_{C_{k}}$ coincide with $m_{k+2} + m_{k+1}$ for each integer $k > 1$.

\paragraph{Off-policy optimization and DPPO masking}
In our setting inference is the main bottleneck. Since all rollouts in a cohort must finish before we can update the policy on that cohort, the longest rollouts largely determine generation time. When inference capacity is available, generating more rollouts in parallel can increase throughput without significantly increasing generation time.

We therefore aim to obtain more policy updates from each generated cohort, amortizing the cost of rollout generation across those updates. At a fixed minibatch size, a larger cohort provides more minibatches and therefore more updates. Alternatively, reducing the minibatch size allows us to perform more updates on the same cohort without generating additional rollouts.

However, performing many off-policy updates leads the target policy to drift from the behaviour policy. As this drift grows, the variance of importance sampling weights increases, making only few tokens dominate the gradient contribution. Hence, the policy gradient estimates become noisier.
To support multiple off-policy updates while limiting this drift, the importance ratio is conventionally clipped to a fixed interval in PPO~\cite{ppo} and GRPO~\cite{grpo},
\[
\operatorname{clip}\!\left(
\frac{\pi_\theta(o_t\mid q,o_{<t})}
     {\mu(o_t\mid q,o_{<t})},
1-\epsilon,
1+\epsilon
\right).
\]
In practice, this clipping is asymmetric in terms of tokens: rare tokens can leave the interval after relatively few updates, whereas already-likely tokens require a larger change. The resulting gradient suppression may penalize exploration more strongly than exploitation and contribute to entropy collapse. Previous works like DPPO~\cite{dppo} and DAPO~\cite{dapo} already discussed these issues.

Instead, we therefore use the DPPO mask. Given the context at token $t$, we can approximate the pointwise divergence between target- and behaviour-policy probabilities as
\[
D_{i,j,t}
=
\left\lvert
\pi_\theta(o_{i,j,t}\mid q_i,o_{i,j,<t})
-
\mu(o_{i,j,t}\mid q_i,o_{i,j,<t})
\right\rvert.
\]
Let $\delta$ be the trust-region radius. The mask is
\[
M_{i,j,t}^{\mathrm{DPPO}}
=
\begin{cases}
0,
& \text{if }(\hat A_{i,j,t}>0\ \land\ r_{i,j,t}>1\ \land\ D_{i,j,t}>\delta) \\
& \phantom{\text{if }}\text{or }(\hat A_{i,j,t}<0\ \land\ r_{i,j,t}<1\ \land\ D_{i,j,t}>\delta), \\[2pt]
1, & \text{otherwise}.
\end{cases}
\]
The mask removes only tokens for which the probability difference exceeds $\delta$ and the gradient update would move the target policy still farther from the behaviour policy.

\paragraph{Credit assignment}\label{par:credit}

Credit assignment governs the variance and specificity of the policy-gradient estimate. Appendix~\ref{sec:reinforce-rft} shows that, with our settings, REINFORCE~\cite{reinforce} without a baseline has the same update direction as online rejection fine-tuning. The main benefit of GRPO- or PPO-style optimization over online RFT~\cite{rft} in this setting is therefore the variance reduction supplied by a baseline. A common choice is the state value $V^\pi(s_t)$; Appendix~\ref{sec:advantages} defines value functions and advantages, and Appendix~\ref{sec:GAE} summarizes the Generalized Advantage Estimation (GAE)~\cite{gae} used by actor--critic PPO with a learned value model $V^\pi_\phi$.

Training a value model adds cost. We therefore use the GRPO group-mean baseline, without reward normalization and following the unnormalized treatment used in Dr.~GRPO~\cite{drgrpo}, for the first two RL runs. Let $q$ be a query, $G$ the group size, and $\{o_i\}_{i=1}^{G}\sim\pi(\cdot\mid q)$ the sampled responses. The advantage assigned to every token in response $o_i$ is
\begin{equation}
\label{eq:group-mean-advantage}
\hat A_{i,t}
=
R_{\lvert o_i\rvert}-\overline R,
\qquad
\overline R
=
\frac{1}{G}\sum_{j=1}^{G}R_{\lvert o_j\rvert}.
\end{equation}
The group-average reward $\overline R$ is a Monte Carlo estimate of the problem-level value, and every token in a rollout receives the same advantage as if all visited states had that value.

\begin{figure}[htbp]
\centering
\includegraphics[width=\linewidth]{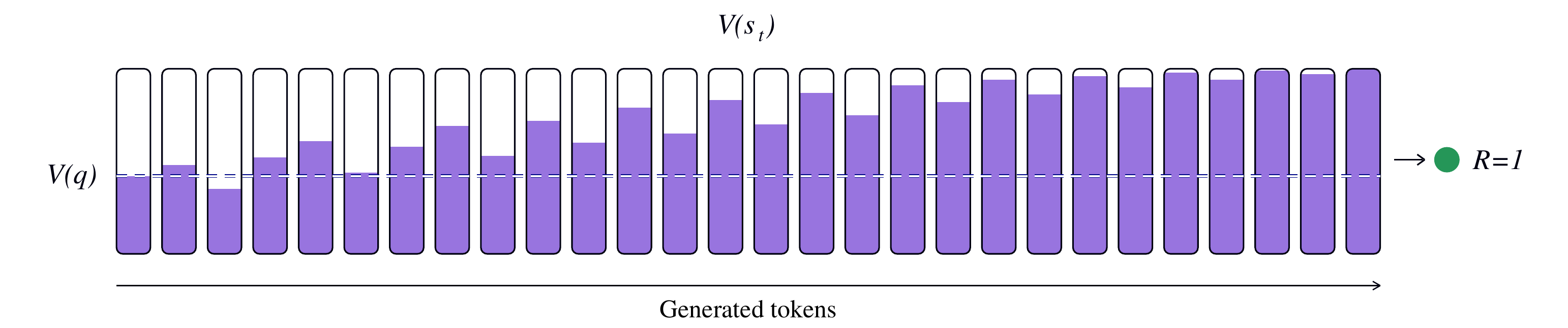}
\caption{Schematic illustration of state values along a successful rollout. The dashed line denotes the problem-level baseline \(V(q)\).}
\label{fig:prefix-values}
\end{figure}

This estimate has two limitations:

\begin{enumerate}
    \item \textbf{Problem value as the baseline}
    \label{item:problem-baseline}
    If state values change substantially over a rollout, the problem-level value may be a poor baseline for many individual states. See the Figure \ref{fig:prefix-values} to see an illustration describing a similar situation.
    \item \textbf{The same advantage for every action}
    \label{item:shared-advantage}
    If only a small number of decisions determine the final outcome, assigning the same advantage to every action provides a diffuse training signal.
\end{enumerate}

The first limitation should matter more for long rollouts, where more visited states may differ substantially from the problem-level value. We hypothesize that the second becomes more important as the policy improves: a failure in an otherwise reliable derivation may stem from only a few decisions, yet the group-level estimator cannot distinguish them from the remaining tokens. These considerations motivate the value model used in the third RL run.

\paragraph{Critic model}\label{par:value-model}
\label{subsubsection:value-model}

Inspired by \citeauthor{justrl2}~\cite{justrl2}, we initialize the value model from the actor and replace the unembedding head with a scalar value head. Let $s_t$ be the rollout prefix through token $t$, $h_t$ the critic's final hidden state for that prefix, and $w$ and $b$ the value-head parameters. The state-value estimate is
\begin{equation}
\label{eq:value-head}
V_\phi(s_t)=w^\top h_t+b.
\end{equation}
Thus, the head is a scalar linear projection without an activation function, as in \parencite{justrl2}. 

We warm up the critic, in two phases, for a total of 501 updates:
\begin{enumerate}
    \item \textbf{First phase} We train the critic for 225 updates
    on already available rollouts previously generated from  older actor policies. In the meantime, we start generating rollouts with the best actor checkpoint until that point, which we use as the starting checkpoint of the third RL run.
    
    \item \textbf{Second phase} Once those rollouts are available,
    we train the critic for another 276 updates on responses
    generated by that checkpoint.
\end{enumerate}

Throughout both phases, the actor remains frozen and only
the critic is trained.

We also apply dynamic sampling~\cite{dapo} to the critic data. In fact, if we included too easy or too hard problems, we would encourage the critic to learn only problem difficulty rather than tracking state values under the actor policy. For stability, we estimate on a large sample the constant output bias that minimizes the initial loss.

For critic training\footnote{See Appendices~\ref{sec:advantages} and ~\ref{sec:GAE} for the definitions of value functions, advantages and GAE.}, we set $\gamma_\mathrm{critic}=\lambda_\mathrm{critic}=1$ . The target for state $s_t$ is therefore the terminal reward of the rollout that continues from that state. We train with the token-averaged mean-squared error
\begin{equation}
\label{eq:critic-loss}
\mathcal{L}_\mathrm{critic}(\phi)
=
\frac{
\displaystyle\sum_{i=1}^{B}\sum_{j=1}^{G}
\sum_{t=1}^{\lvert o_{i,j}\rvert}
\left(V_\phi(s_{i,j,t})-R_{i,j}\right)^2
}{
\displaystyle 2\sum_{i=1}^{B}\sum_{j=1}^{G}
\lvert o_{i,j}\rvert
}.
\end{equation}

As shown in the warm-up curves in Figure~\ref{fig:value-model-warmup}, the second phase converges within a few updates. Nevertheless, the critic reduces downstream policy-gradient variance, which helps stabilize training, so we extend the second phase beyond what would otherwise appear to be a more natural stopping point. We refer readers to Section~\ref{par:third-rl-run} for the description of the third RL run, in which we use the critic.

\plate[1.0]{%
  \includegraphics[width=\linewidth]{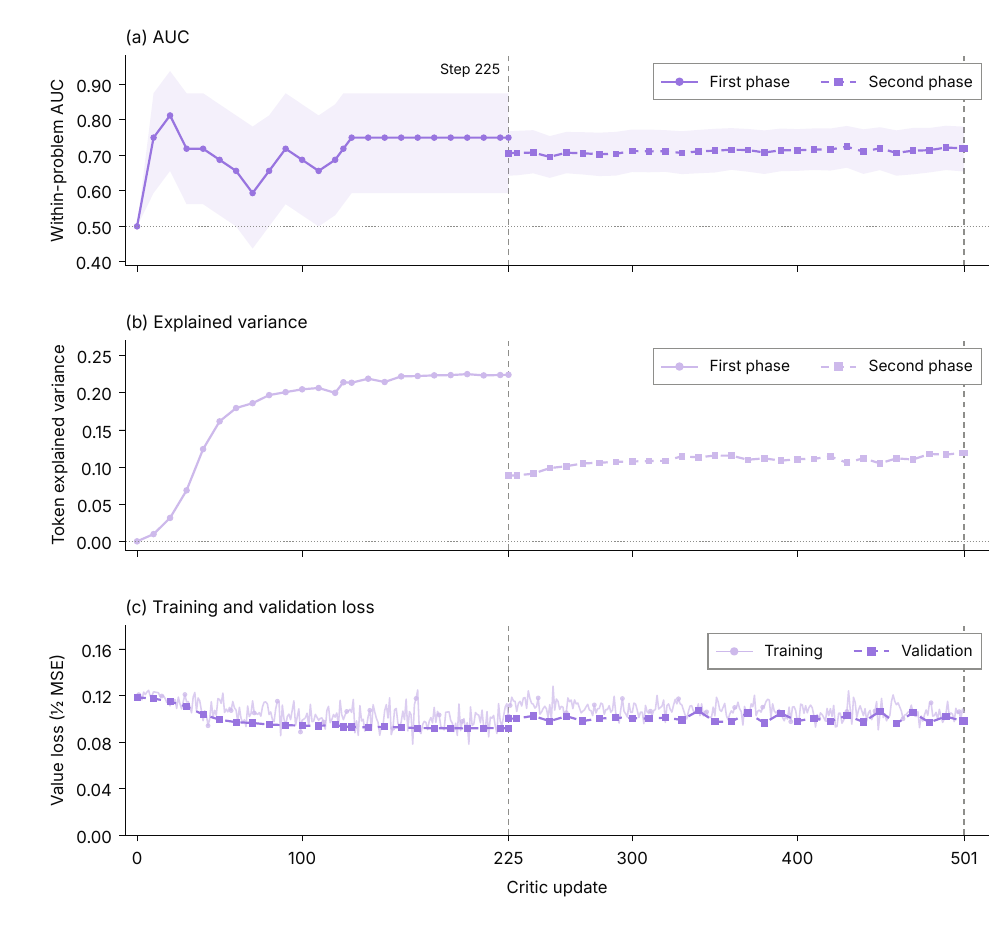}
}{Value-model metrics during offline warm-up:
(a) AUC on validation subset, (b) explained variance on validation subset, and
(c) training and validation loss. All the plots share the x-axis that tracks the number of critic updates. These metrics are useful to evaluate the critic performance. AUC tells us about how able the critic is at ranking correct rollouts above incorrect ones, whereas explained variance measures the fraction of return variance removed by using the critic as a baseline. The dashed lines mark the phase transition at
update 225 and the end of warm-up at update 501.
\label{fig:value-model-warmup}}

\subsection{RL Training Sequence and Results}\label{subsec:training-sequence}\label{subsubsec:experiments}

We perform three DPPO runs, all with AdamW for both actor and critic (present only in the third run). Throughout, we set the DPPO trust-region radius to $\delta=0.15$, and clip the importance-sampling weight at 5.

\begin{table}[!htbp]
\centering
\caption{Benchmark accuracy (\%) across post-training stages.
Off-policy distillation, DPO pre-RL, First RL, and Second RL were
evaluated with a context limit of 65,536 and a response-token cap
of 61,440. Columns marked \(\dagger\) (SFT post-RL and Third RL) were evaluated with a
context limit of 131,072 and a response-token cap of 126,976,
following the decision to extend the model's context length and
train for longer reasoning. Gains are percentage-point differences
from the preceding stage, computed from displayed scores. Detailed results at 65K and 131K context lengths are reported in Appendix~\ref{app:context-length-results}.}
\label{tab:accuracy-131072}

\pgcaptionfont
\newcommand{\stagegain}[1]{\textcolor{pgtwilight}{#1}}
\setlength{\tabcolsep}{4pt}
\renewcommand{\arraystretch}{1.15}
\begin{tabularx}{\linewidth}{@{}Xr*{5}{rr}@{}}
\toprule
Benchmark &
\multicolumn{1}{c}{\shortstack{Off-policy\\distillation}} &
\multicolumn{2}{c}{\shortstack{DPO\\pre-RL}} &
\multicolumn{2}{c}{\shortstack{First\\RL}} &
\multicolumn{2}{c}{\shortstack{Second\\RL}} &
\multicolumn{2}{c}{\shortstack{SFT\\post-RL\textsuperscript{\(\dagger\)}}} &
\multicolumn{2}{c@{}}{\shortstack{Third\\RL\textsuperscript{\(\dagger\)}}} \\
\cmidrule(lr){2-2}\cmidrule(lr){3-4}\cmidrule(lr){5-6}
\cmidrule(lr){7-8}\cmidrule(lr){9-10}\cmidrule(l){11-12}
& Score & Score & Gain & Score & Gain & Score & Gain
& Score & Gain & Score & Gain \\
\midrule[0.9pt]
AIME 2025
& 77.08 & 80.52 & \stagegain{+3.44}
& 85.62 & \stagegain{+5.10}
& 87.71 & \stagegain{+2.09}
& 89.17 & \stagegain{+1.46}
& 90.21 & \stagegain{+1.04} \\

AIME 2026
& 81.67 & 83.54 & \stagegain{+1.87}
& 87.50 & \stagegain{+3.96}
& 87.40 & \stagegain{$-$0.10}
& 91.35 & \stagegain{+3.95}
& 94.01 & \stagegain{+2.66} \\

HMMT Feb.\ 2025
& 70.42 & 76.15 & \stagegain{+5.73}
& 80.73 & \stagegain{+4.58}
& 80.62 & \stagegain{$-$0.11}
& 89.69 & \stagegain{+9.07}
& 91.35 & \stagegain{+1.66} \\

HMMT Feb.\ 2026
& 64.68 & 62.41 & \stagegain{$-$2.27}
& 73.01 & \stagegain{+10.60}
& 72.25 & \stagegain{$-$0.76}
& 80.78 & \stagegain{+8.53}
& 83.62 & \stagegain{+2.84} \\

APEX Shortlist
& 23.27 & 21.81 & \stagegain{$-$1.46}
& 33.44 & \stagegain{+11.63}
& 32.05 & \stagegain{$-$1.39}
& 43.42 & \stagegain{+11.37}
& 50.80 & \stagegain{+7.38} \\

BeyondAIME
& 57.16 & 52.25 & \stagegain{$-$4.91}
& 65.28 & \stagegain{+13.03}
& 65.34 & \stagegain{+0.06}
& 72.38 & \stagegain{+7.04}
& 74.25 & \stagegain{+1.87} \\

ArXivMath May 2026
& 20.08 & 15.39 & \stagegain{$-$4.69}
& 25.08 & \stagegain{+9.69}
& 24.14 & \stagegain{$-$0.94}
& 23.52 & \stagegain{$-$0.62}
& 25.08 & \stagegain{+1.56} \\
\bottomrule
\end{tabularx}
\end{table}

\begin{table}[!htbp]
\centering
\caption{Mean generated response length (tokens) at a context limit of 131,072 and a response-token cap of 126,976.}
\label{tab:rollout-lengths-131072}

\pgcaptionfont
\setlength{\tabcolsep}{5pt}
\renewcommand{\arraystretch}{1.15}
\begin{tabularx}{\linewidth}{@{}Xrrrrrr@{}}
\toprule
Benchmark & \shortstack{Off-policy\\distillation} & \shortstack{DPO\\pre-RL} & \shortstack{First\\RL} & \shortstack{Second\\RL} & \shortstack{SFT\\post-RL} & \shortstack{Third\\RL} \\
\midrule[0.9pt]
AIME 2025 & 27,725 & 35,506 & 22,736 & 21,226 & 23,474 & 31,968 \\
AIME 2026 & 24,219 & 32,800 & 19,502 & 18,376 & 21,121 & 29,863 \\
HMMT Feb.\ 2025 & 35,037 & 42,709 & 26,973 & 24,393 & 28,309 & 39,647 \\
HMMT Feb.\ 2026 & 37,825 & 44,842 & 28,404 & 26,815 & 33,042 & 44,297 \\
APEX Shortlist & 50,485 & 59,860 & 38,747 & 32,359 & 54,239 & 75,007 \\
BeyondAIME & 39,323 & 49,517 & 30,540 & 27,404 & 36,936 & 52,387 \\
ArXivMath May 2026 & 40,908 & 65,128 & 34,097 & 29,474 & 46,560 & 68,002 \\
\bottomrule
\end{tabularx}
\end{table}

\paragraph{First RL run}

The first run tests whether a high learning rate and a large number of off-policy updates can produce rapid gains. By update 213, we found a good balance of performance and brevity as shown in Table \ref{tab:accuracy-131072} and Table \ref{tab:rollout-lengths-131072}. Beyond that, excessive drift between target and behaviour policies produces a sharper optimization regime and eventual reward collapse, so we select checkpoint 213. Each cohort contains 704 problems with 8 rollouts each. After discarding non-informative groups, we split the remainder into minibatches of 32 problems; because the retained count varies by cohort, so does the number of optimizer updates. Responses are capped at 61,440 tokens despite the preceding model's exposure to longer traces, both to encourage concise reasoning and to reduce rollout time. \enlargethispage{\baselineskip}We perform 19 off-policy updates per behaviour policy on average at a constant learning rate of~${1.5\times10^{-5}}$.

\plate[1.0]{%
  \includegraphics[width=\linewidth]{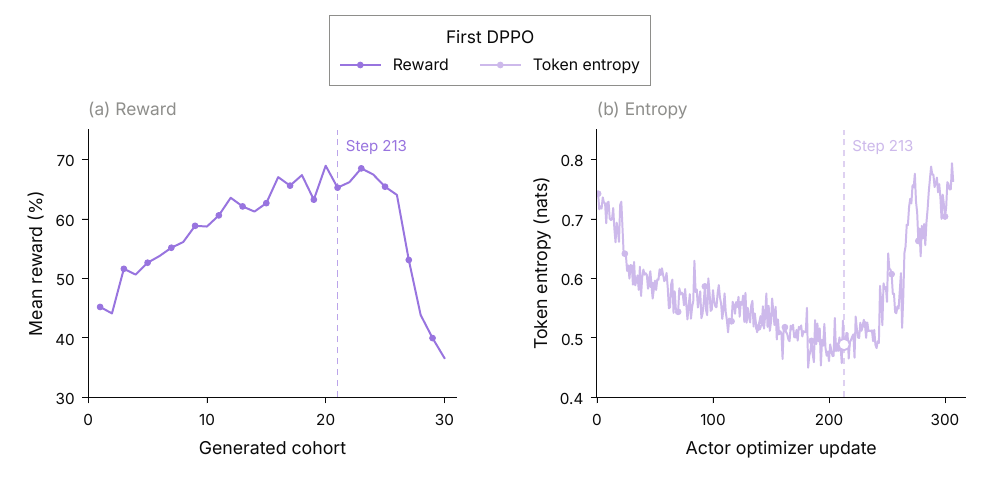}%
}{First DPPO: complete training trajectory, including continuation beyond the selected checkpoint. Left: mean binary correctness reward across all responses in complete generated cohorts, including truncated responses. Right: mean token entropy on informative training minibatches. Dashed lines mark checkpoint 213.
\label{fig:first-dppo}}

\paragraph{Second RL run}

The second run starts from the first run's selected checkpoint. To improve stability, we double the minibatch size, thereby reducing the number of off-policy updates, and add data to broaden the training signal. More details about the data we added are provided in the paragraph \hyperref[par:rl-problem-generation]{\textit{RL problem generation}} in Section~\ref{par:rl-problem-generation}. Performance improves slightly by update 179, after which target--behaviour divergence again leads to reward collapse; we therefore select checkpoint 179. Cohort and group sizes remain 704 problems and 8 rollouts, while the minibatch increases to 64 problems thereby reducing the mean number of off-policy updates per behaviour policy from 19 to 9. We begin at $1\times10^{-5}$ and, because reward improves too slowly, increase the learning rate to $2\times10^{-5}$ after 73 actor updates. We also reduce the gradient-norm clipping threshold to $0.02$ to limit spikes that could drive the target policy too far from the behaviour policy. The response cap remains 61,440 tokens.

\plate[1.0]{%
  \includegraphics[width=\linewidth]{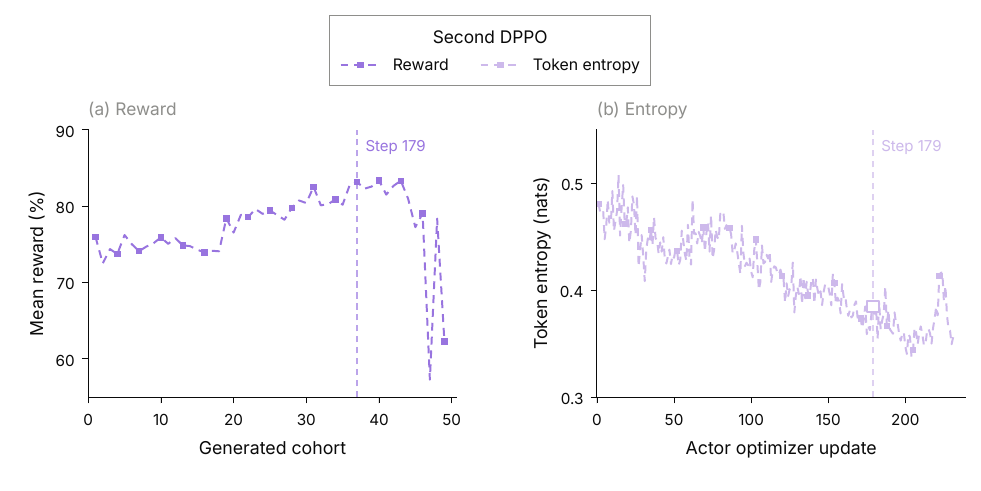}%
}{Second RL run: complete training trajectory, including continuation beyond the selected checkpoint. Left: mean binary correctness reward across all responses in complete generated cohorts, including truncated responses. Right: mean token entropy on informative training minibatches. Dashed lines mark checkpoint 179.
\label{fig:second-dppo}}

\paragraph{Intermediate SFT and model soup}
The second run's limited improvement and eventual instability suggested that further RL on the same trajectory would be inefficient. As we discuss in the paragraph \hyperref[par:sft-rl]{\textit{Interleaving SFT and RL}} in Section~\ref{par:sft-rl}, training through SFT is more efficient in these cases. We therefore perform two short SFT stages on the targeted mixture described above, averaging checkpoints once between them and again after the second stage. The resulting soup initializes the final RL run. This stage also expands the supported context length from 65k to 131k tokens; its benchmark effect appears in the SFT post-RL column of Table~\ref{tab:accuracy-131072}.

\paragraph{Third RL run}\label{par:third-rl-run}
The third run starts from the post-SFT soup and uses the learned critic for token-dependent advantages. The model improves substantially within 59 actor updates. We attribute this to the critic together with the stronger initialization produced by targeted SFT and checkpoint averaging, rather than to the critic alone. For the actor, we set $\gamma_\mathrm{policy}=1$ and
\[
\lambda_\mathrm{policy}=0.5^{1/L},
\]
where $L$ is the rollout length. Under the indexing used in our implementation, this choice is intended to keep the contribution of the terminal reward to the earliest token constant with respect to $L$:
\[
\lambda_\mathrm{policy}^{L}=0.5.
\]
For the critic, we use $\gamma_\mathrm{critic}=\lambda_\mathrm{critic}=1$, as described above.

We increase the response cap to 126,976 tokens to train long-context reasoning. The cohort contains 256 problems, the minibatch contains 32 problems, and we perform 7 off-policy updates per behaviour policy on average. The actor uses a constant learning rate of $1\times10^{-5}$. For every actor update, the critic receives two updates at a learning rate of $5\times10^{-6}$ so that it can track the changing policy.

The policy objective is otherwise the same as in the first two runs, but the advantages are estimated by the critic rather than by the group-mean baseline. Appendix~\ref{sec:GAE} gives the corresponding GAE definition.

\plate[1.0]{%
  \includegraphics[width=\linewidth]{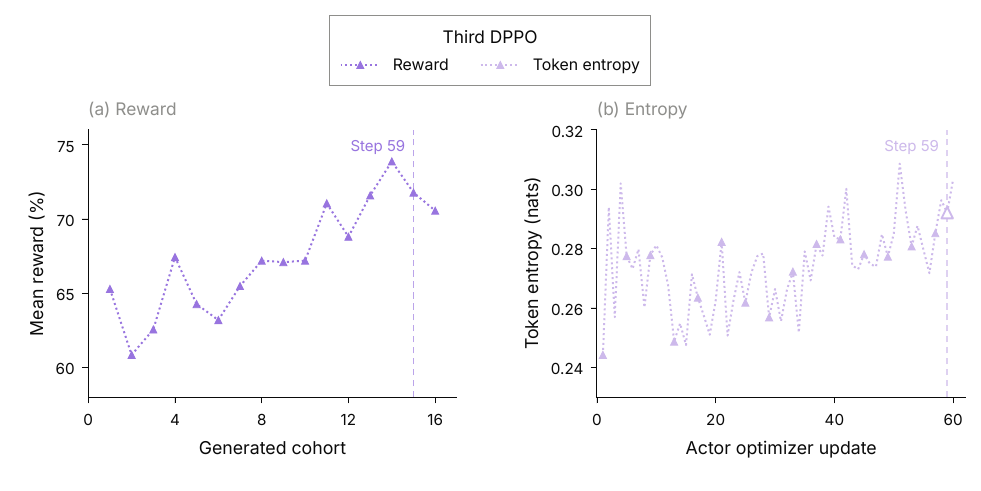}%
}{Third RL run: complete training trajectory, including continuation beyond the selected checkpoint. Left: mean binary correctness reward across all responses in complete generated cohorts, including truncated responses. Right: mean token entropy on informative training minibatches. Dashed lines mark checkpoint 59.
\label{fig:third-dppo}}

\section{Evaluation}\label{sec:evaluation}
\subjectline{Evaluation}{Mathematical reasoning}

We evaluated Limite on seven mathematical reasoning benchmarks: AIME 2025 and 2026, HMMT February 2025 and 2026, the MathArena Apex Shortlist, BeyondAIME, and ArXivMath (May 2026). This suite combines recurring competitions, harder olympiad collections, and research-level problems from recent arXiv papers, while keeping a single property across all of them: every answer can be extracted and verified automatically. 

This requirement allows every completion to be graded by deterministic, rule-based comparison against the reference answer rather than by an LLM judge. More importantly, it allows each completion to be scored under increasingly tolerant answer-extraction rules, ranging from a \textit{strict} rule that accepts only a well-formed final \verb|\boxed{}| answer, through a \textit{lenient} rule that recovers plausible final answers written in the wrong form, to a \textit{permissive} rule that accepts the correct value anywhere in the completion. We also track the gaps between these three scores, together with the rates of truncated and malformed outputs. \\
Across pre-training, supervised fine-tuning and reinforcement learning, these signals show how much of a change in performance comes from genuine mathematical ability and how much comes from a change in output format. \\
For the headline scores reported in Table~\ref{tab:math-evaluation}, we rely on official graders wherever they exist. For AIME 2025 and 2026, HMMT February 2025 and 2026, and the Apex Shortlist, completions are scored with MathArena's published grader. BeyondAIME does not release an executable grader, and for ArXivMath we found that the MathArena grader produced both false-positive and false-negative verdicts; for these two benchmarks we therefore extract the final \verb|\boxed{}| answer and compare it with the reference using Math-Verify~\cite{mathverify}.
We release our evaluation suite implementation at \mbox{\href{https://github.com/paradigma-inc/limite-violetto}{\nolinkurl{github.com/paradigma-inc/limite-violetto}}}.

\begin{table}[!htbp]
\centering
\pgcaptionfont
\setlength{\tabcolsep}{1.8pt}
\caption{Results on mathematical reasoning benchmarks (pass@1, \%, averaged over
32 samples per problem). Models above the line have at most 9B total
parameters, and \textbf{bold} marks the best score in this group, together with
any score whose standard-error interval overlaps it (Table~\ref{tab:math-evaluation-sem}). Models below
the line are larger models included for reference. Unless marked, all scores
come from our own evaluation: AIME, HMMT and the Apex Shortlist are graded with
MathArena's published grader, while BeyondAIME and ArXivMath are graded with
Math-Verify on the final \texttt{\textbackslash boxed\{\}} answer.
\(\dagger\)~Score taken from the MathArena leaderboard or from the official
model card instead of being reproduced by us.}\label{tab:math-evaluation}
\begin{tabularx}{\linewidth}{@{}>{\centering\arraybackslash}Xccccccccc@{}}
\toprule
Model & Release & \shortstack[c]{Params.\\total} & \shortstack[c]{AIME\\2026} &
\shortstack[c]{HMMT\\Feb.\ 2026} & \shortstack[c]{APEX\\Shortlist} &
\shortstack[c]{Beyond\\AIME} & \shortstack[c]{AIME\\2025} &
\shortstack[c]{HMMT\\Feb.\ 2025} &
\shortstack[c]{ArXivMath\\May 2026} \\
\midrule
\rowcolor{pgtwilight!22!white}
\textbf{Limite 1B - Violetto}
                  & \textbf{2026-09} & \textbf{1B} & \textbf{94.01} & \textbf{83.62} & \textbf{50.80} & \textbf{74.25} & 90.21 & \textbf{91.35} & 25.08 \\
VibeThinker-1.5B
                  & 2025-11 & 1.5B & 70.94 & 47.25 & 10.84 & 48.06 & 72.92 & 50.52 & 18.13 \\
MiniCPM5-2B
                  & 2026-09 & 2B & 90.21 & 67.80 & 26.86 & 60.59 & 86.98 & 78.02 & 21.80 \\
VibeThinker-3B
                  & 2026-06 & 3B & \textbf{93.85} & 78.98 & 46.41 & 72.00 & \textbf{92.19} & 88.23 & \textbf{29.77} \\
Qwen3.5-4B
                  & 2026-02 & 4B & \phantom{\raisebox{0.35ex}{\(\dagger\)}}\,89.66\,\raisebox{0.35ex}{\(\dagger\)} & \phantom{\raisebox{0.35ex}{\(\dagger\)}}\,72.86\,\raisebox{0.35ex}{\(\dagger\)} & \phantom{\raisebox{0.35ex}{\(\dagger\)}}\,32.51\,\raisebox{0.35ex}{\(\dagger\)} & 61.25 & 82.19 & 74.58 & 26.33 \\
Olmo-3-7B-Think
                  & 2025-11 & 7B & 69.38 & 41.95 & 13.16 & 41.00 & 68.02 & 42.60 & 18.05 \\
Olmo-3-7B-Instruct
                  & 2025-11 & 7B & 44.27 & 31.91 & 6.18 & 22.81 & 39.38 & 27.92 & 11.33 \\
OLMo-Hybrid-Think-SFT-7B
                  & 2026-02 & 7B & 62.08 & 35.89 & 10.57 & 36.41 & 53.65 & 35.10 & 15.47 \\
Qwen3.5-9B
                  & 2026-02 & 9B & 90.42 & 70.36 & 30.72 & 65.56 & 88.75 & 83.02 & \textbf{31.64} \\
\midrule
\addlinespace[2pt]
Qwen3.5-27B\raisebox{0.35ex}{\(\dagger\)}
                  & 2026-02 & 27B & 91.67 & 81.06 & 52.13 & --- & --- & --- & --- \\
Qwen3.6-27B\raisebox{0.35ex}{\(\dagger\)}
                  & 2026-04 & 27B & 94.10 & 84.30 & --- & --- & --- & 93.80 & --- \\
MUSE-Glimmer-30B
                  & 2026-08 & 30B & 93.44 & 80.87 & 51.66 & 70.00 & 92.71 & 90.21 & 28.91 \\
Gemma-4-31B-IT
                  & 2026-03 & 31B & 89.17 & 77.08 & 40.16 & 71.00 & 87.81 & 86.88 & 35.78 \\
Qwen3.6-35B-A3B\raisebox{0.35ex}{\(\dagger\)}
                  & 2026-04 & 35B & 92.70 & 83.60 & --- & --- & --- & 90.70 & --- \\
NVIDIA-Nemotron-3-Super\raisebox{0.35ex}{\(\dagger\)}
                  & 2026-03 & 120B & 91.67 & 84.85 & 57.45 & --- & --- & --- & --- \\
GLM 5.2\raisebox{0.35ex}{\(\dagger\)}
                  & 2026-06 & 753B & 99.20 & 92.50 & 68.09 & --- & --- & --- & 56.67 \\
Gemini 3 Pro (preview)\raisebox{0.35ex}{\(\dagger\)}
                  & 2025-11 & n.d. & 91.67 & 86.36 & 66.49 & --- & 95.00 & 97.50 & --- \\
\bottomrule
\end{tabularx}
\end{table}

\section{Conclusion}\label{sec:conclusion}
\subjectline{Conclusion}{Efficiency, limits, and next steps}

We presented Limite 1B - Violetto, a model trained from scratch with the objective of maximizing mathematical capability per parameter. We oriented our decisions across the entire stack, from architecture, to data curation, to optimization and post-training around the goal of maximizing sample efficiency. A mathematics-focused pretraining corpus provided the foundation for subsequent distillation, targeted supervised fine-tuning, and reinforcement learning with verifiable rewards. The resulting one-billion-parameter model is competitive with substantially larger models on hard mathematical problems. Its compact size also supports efficient inference, although its deep architecture can limit decoding speed and the model has not been fully optimized for inference.

This specialization has clear limitations. Limite is only lightly instruction-tuned, and its mathematical performance does not imply comparable capability in general instruction following or other domains. In broader research workflows, it could serve as a mathematical reasoning component within multi-agent systems, with complementary models and tools providing coordination, capabilities outside mathematics, and verification.

Limite 1B - Violetto is our first step towards building models for autonomous science, and will act as a concrete base for our upcoming work. We are already developing the next models and systems around the objective that motivates this effort: maximizing discovery per unit of computation.

\par\vspace{2\baselineskip}
\noindent\makebox[\linewidth][c]{\includegraphics[width=\linewidth,
  height=\dimexpr\pagegoal-\pagetotal-3\baselineskip\relax,
  keepaspectratio]{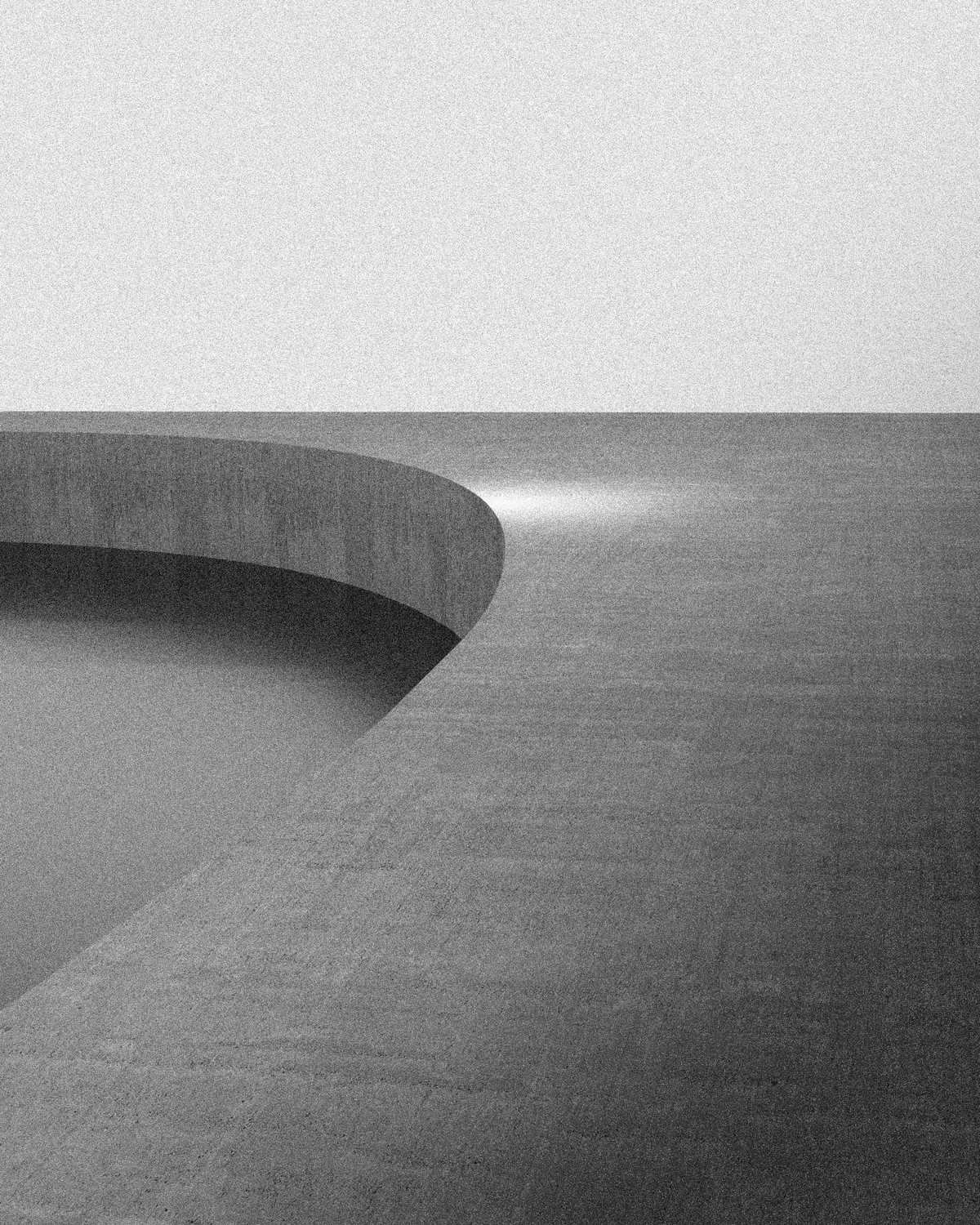}}

\subjectline{References}{Numbered notes}
\filenote{}
\printbibliography[title={References}]

\section*{Appendix}\label{sec:appendix}
\subjectline{Appendix}

\appendix

\section{NorMuon}
Algorithm~\ref{alg:normuon} gives the NorMuon update for one hidden weight
matrix in the final pre-training run. Data-parallel communication is omitted:
gradients are averaged with a reduce-scatter along the bank dimension, so every
rank updates complete matrices, and the result is all-gathered. The
coefficients $(a_k, b_k, c_k)$ are the Polar Express schedule for 5 iterations
with safety factor $2\cdot10^{-2}$ and cushion 2. Steps 2 and 4, in violet, are
where the update departs from the original NorMuon~\cite{li2025normuon}.
\begin{algorithm}[H]
\caption{NorMuon update for one hidden weight matrix}
\label{alg:normuon}
\small
\begin{algorithmic}[1]
\Require weight $W \in \mathbb{R}^{m \times n}$ (fp32 master), gradient $G_t$ averaged over ranks,
         momentum buffer $B_{t-1}$, second-moment buffer $\nu_{t-1}$ (both fp32, $B_0 = 0$, $\nu_0 = 0$)
\Require momentum $\mu_t$, $\beta_2 = 0.9$, learning rate $\eta_t$, weight decay $\lambda = 1.2$,
         Polar Express coefficients $(a_k, b_k, c_k)_{k=1}^{5}$
\Statex
\Statex \textbf{1. Nesterov momentum} \hfill (fp32)
\State $B_t \gets \mu_t B_{t-1} + (1-\mu_t)\, G_t$
\State $\tilde{G}_t \gets (1-\mu_t)\, G_t + \mu_t B_t$
\Statex
\Statex \color{pgtwilight}\textbf{2. Polar Express orthogonalisation} \hfill (bf16, fp32 accumulation)
\State $X_0 \gets \operatorname{bf16}(\tilde{G}_t)$
\State $X_0 \gets X_0 \,/\, \big( 1.02\, \lVert X_0 \rVert_F + 10^{-6} \big)$
  \Comment{$\lVert X_0 \rVert_2 \le \lVert X_0 \rVert_F$, so spectral norm $< 1$}
\For{$k = 1, \dots, 5$}
  \If{$m > n$}
    \State $A \gets X_{k-1}^{\top} X_{k-1}$;\quad
           $X_k \gets a_k X_{k-1} + X_{k-1}\big(b_k A + c_k A^2\big)$
  \Else
    \State $A \gets X_{k-1} X_{k-1}^{\top}$;\quad
           $X_k \gets a_k X_{k-1} + \big(b_k A + c_k A^2\big) X_{k-1}$
  \EndIf
\EndFor
\State $U \gets X_5$
\Statex
\Statex \color{pgink}\textbf{3. NorMuon variance reduction} \hfill (statistics in fp32)
\State $r \gets$ columns if $m \ge n$, otherwise rows
  \Comment{the longer dimension; $W_v$ always uses columns}
\State $s \gets \operatorname{mean}_r (U \odot U)$
  \Comment{$s \in \mathbb{R}^{m}$ or $\mathbb{R}^{n}$; $n_r = $ size of $r$}
\State $\rho \gets \big( n_r \textstyle\sum s \big)^{1/2}$
  \Comment{$= \lVert U \rVert_F$}
\State $\nu_t \gets \beta_2\, \nu_{t-1} + (1 - \beta_2)\, s$
\State $\sigma \gets \max(\nu_t,\, 10^{-10})^{-1/2}$
\State $\rho' \gets \big( n_r \textstyle\sum s \odot \sigma^2 \big)^{1/2}$
  \Comment{$= \lVert U \odot \sigma \rVert_F$}
\State $\Delta \gets U \odot \sigma \cdot \rho \,/\, \max(\rho',\, 10^{-10})$
  \Comment{$\sigma$ broadcast along $r$; $\lVert \Delta \rVert_F = \lVert U \rVert_F$}
\Statex
\Statex \color{pgtwilight}\textbf{4. Cautious weight decay and update} \hfill (fp32 master)
\State $C \gets \mathbb{1}\big[\, \Delta \odot W > 0 \,\big]$
  \Comment{strict: no decay where $\Delta_{ij} = 0$}
\State $W \gets W - \lambda\, \eta_t^2\, (C \odot W) - \eta_t\, \Delta$
\State \Return $W$, $B_t$, $\nu_t$
\end{algorithmic}
\end{algorithm}

\section{REINFORCE, without any baseline, is actually RFT}\label{sec:reinforce-rft}

$ 
\text{Let } o_i \text{ be the response generated by the policy $\pi_\theta$ to a query } q_i \text{, then:}
$
\begin{equation}
\label{eq:reward}
R_{i,t} = \begin{cases}
    1& \quad \text{if } t = \lvert o_i \rvert \text{ and } o_i \text{ has a correct final answer} \\
    0& \quad \text{otherwise}
\end{cases}   
\end{equation}

Let $B$, $C$ and $\gamma$ be, respectively, the batch, its subset containing only correct rollouts and the discount factor, which controls how much a future reward is valued compared with an immediate one.
Then, the policy gradient estimate in the REINFORCE algorithm is:

\begin{align}
    \hat{g}_{\mathrm{REINFORCE}} &= \frac{1}{\lvert B \rvert}\sum_{i=1}^{\lvert B \rvert}\left[ \sum_{t=1}^{\lvert o_i \rvert} \gamma^t \nabla_{\theta} \log\pi_\theta(o_{i,t} \lvert q_i, o_{i,<t}) \sum_{\tau =t}^{\lvert o_i \rvert} \gamma^{\tau - t}R_{i,\tau}  \right] \notag\\
    &= \frac{1}{\lvert B \rvert}\sum_{i=1}^{\lvert B \rvert}\left[ \sum_{t=1}^{\lvert o_i \rvert} \nabla_{\theta} \log\pi_\theta(o_{i,t} \lvert q_i, o_{i,<t}) \sum_{\tau =t}^{\lvert o_i \rvert} R_{i,\tau}  \right] \tag{i}\\
    &= \frac{1}{\lvert B \rvert}\sum_{i=1}^{\lvert B \rvert} R_{i,\lvert o_i \rvert} \sum_{t=1}^{\lvert o_i \rvert} \nabla_{\theta} \log\pi_\theta(o_{i,t} \lvert q_i, o_{i,<t}) \tag{ii}\\
    &= \frac{1}{\lvert B \rvert} \sum_{i \in C}\sum_{t=1}^{\lvert o_i \rvert} \nabla_{\theta} \log\pi_\theta(o_{i,t} \lvert q_i, o_{i,<t}) \tag{iii}
\end{align}

The equalities follow from:

\begin{enumerate}
    \item[(i)] Setting $\gamma = 1$
    \item[(ii)] Assigning zero reward to each state different from the last one
    \item[(iii)] Assigning $R_{i,\lvert o_i \rvert} = 0$ to wrong rollouts and $R_{i,\lvert o_i \rvert} = 1$ to correct ones 
\end{enumerate}

As you can see, under these settings, REINFORCE and online RFT have the same update direction. Indeed, the policy gradient is proportional to the negative gradient of the online RFT loss, which is the cross entropy loss on correct solutions sampled from $\pi_\theta$.

Formally:

\begin{align}
    \mathcal{L}_\mathrm{RFT}(\theta) &= - \frac{1}{\lvert C \rvert}\sum_{i \in C}\sum_{t=1}^{\lvert o_i \rvert} \log\pi_\theta(o_{i,t} \lvert q_i, o_{i,<t}) \\
    \hat{g}_\mathrm{RFT} &= - \frac{1}{\lvert C \rvert}\sum_{i \in C}\sum_{t=1}^{\lvert o_i \rvert} \nabla_{\theta} \log\pi_\theta(o_{i,t} \lvert q_i, o_{i,<t})
\end{align}

Therefore:
$$
\hat{g}_{\mathrm{REINFORCE}} \propto - \hat{g}_\mathrm{RFT}
$$

\section{Value functions and advantages definitions}\label{sec:advantages}

\begin{itemize}
    \item $Q^\pi(s_t,a_t)$ is the expected return of taking the action $a_t$ in the state $s_t$ and then following policy $\pi$.
    \item $V^\pi(s_t)$ is the expected return from the state $s_t$ under policy $\pi$.
    \item $A^\pi(s_t,a_t)$ is the advantage of taking the action $a_t$ in the state $s_t$ for the policy $\pi$
\end{itemize}

Namely:

\begin{align}
    Q^\pi(s,a) &= \mathbb{E}_\pi \left[\sum_{\tau=0}^\infty\gamma^\tau R_{i,t + \tau} \;\middle|\;  s_t = s, a_t = a \right] \\
    V^\pi(s) &= \mathbb{E}_{a \sim \pi( \cdot \mid s)} \left[ Q^\pi(s,a) \right] \\
    A^\pi(s,a) &= Q^\pi(s,a) - V^\pi(s)
\end{align}

In our case, $\gamma=1$ and rewards are defined as in Equation~\eqref{eq:reward}.

Let $s$ be $[q, o_{<t}]$, where $q$ is a query and $o$ is the rollout sampled by $\pi$ as a response to $q$, then under our settings:

\begin{align}
    Q^\pi(s,o_t) &= \mathbb{E}_\pi \left[ R_{\lvert o \rvert} \;\middle|\;  s_t = s, a_t = o_{t} \right] \\
    V^\pi(s) &= \mathbb{E}_\pi \left[ R_{\lvert o \rvert} \mid s_t = s \right] \\
    A^\pi(s,o_t) &= Q^\pi(s,o_{t}) - V^\pi(s)
\end{align}

\section{GAE}\label{sec:GAE}

\label{eq:gae-lambda}
GAE allows us to control the bias-variance trade-off through $\lambda \in [0,1]$: lower values imply more bootstrapping from the value model, while $\lambda = 1$ uses an unbiased Monte Carlo estimate of the return.

In detail, given the timestep $t=1,\dots, T$, set the value after the final timestep equal to $0$: $V^\pi_\phi(s_{T+1})=0$, then:

\begin{align}
    \delta_{i,t}
    =& R_{i,t}+\gamma V^\pi_\phi(s_{t+1})-V^\pi_\phi(s_t)
    \qquad \\
    \hat A^{\mathrm{GAE}}_{i,t}
    =& \sum_{\tau=0}^{T-t}(\gamma\lambda)^\tau\delta_{i,t+\tau}
\end{align}

In our case, $T=\lvert o_i \rvert$ which is the length of the model rollout $o_i$, $\gamma=1$, and rewards are defined as in
Equation~\eqref{eq:reward}.

Therefore:

\begin{align}
    \delta_{i,t} &= \begin{cases}
        R_{i,\lvert o_i \rvert} - V^\pi_\phi(s_{\lvert o_i \rvert}) \qquad &\text{if } t = \lvert o_i \rvert   \\
        V^\pi_\phi(s_{t+1}) - V^\pi_\phi(s_{t}) \qquad &\text{otherwise}
    \end{cases} \\
    \hat{A}^\mathrm{GAE}_{i,t} &= \sum_{\tau = 0}^{\lvert o_i \rvert - t } \lambda^\tau \delta_{i,t+\tau} \\
    \hat{A}^\mathrm{GAE}_{i,t} &= \sum_{\tau = 0}^{\lvert o_i \rvert - t} \delta_{i,t+\tau} = R_{i,\lvert o_i \rvert} - V^\pi_\phi(s_t) \qquad \text{if } \lambda =1 
\end{align}

\section{Length-normalized DPO objective}
\label{app:dpo}

Let $\pi_\theta$ be the trained policy and $\pi_\mathrm{ref}$ the frozen
reference policy. Both are initialized from the distilled soup. Each
training example is a query $q$ with a preferred response $o^{+}$ and a
dispreferred response $o^{-}$. Standard
DPO~\cite{rafailov2024directpreferenceoptimizationlanguage} defines the
implicit reward as the \emph{summed} log-ratio over the response tokens. We
instead use its \emph{mean} over the response
length~\cite{ahrabian2025practicalanalysishumanalignment}:
\begin{equation}
\label{eq:dpo-reward}
r_\theta(q,o)
=
\frac{1}{\lvert o\rvert}
\sum_{t=1}^{\lvert o\rvert}
\log
\frac{\pi_\theta(o_t\mid q,o_{<t})}
     {\pi_\mathrm{ref}(o_t\mid q,o_{<t})}.
\end{equation}
The loss over the preference dataset $\mathcal{D}$ is
\begin{equation}
\label{eq:dpo-loss}
\mathcal{L}_\mathrm{DPO}(\theta)
=
-\,\mathbb{E}_{(q,o^{+},o^{-})\sim\mathcal{D}}
\Big[
\log\sigma\!\Big(
\beta\big(r_\theta(q,o^{+})-r_\theta(q,o^{-})\big)
\Big)
\Big],
\end{equation}
with $\beta=5$. With the summed reward, the margin grows with the number of
tokens. A pair whose responses differ in length therefore gets a reward gap
that partly reflects length rather than quality. Dividing by $\lvert o\rvert$
puts the rewards of both responses on a per-token scale, which removes this
bias.

\section{Compute estimates for Figure~\ref{fig:compute}}
\label{app:compute}

Training compute in Figure~\ref{fig:compute} is the analytical $6ND$
estimate, with $N$ the parameter count (active parameters for
mixture-of-experts models) and $D$ the number of training tokens. It is an
estimate, not a measured total. For Limite 1B - Violetto it counts
pretraining, offline distillation and SFT, and excludes RL. For the other
models it comes from public reports and model cards, and the training stages
counted differ by model.

The marked points need the following qualifications.
\begin{itemize}
  \item $^{*}$VibeThinker 3B: the Coder training stage only.
  \item $^{\dagger}$MiniCPM5 2B: the 1T midtraining tokens only. The solid
    arrow marks the earlier pretraining this leaves out.
  \item $^{\ddagger}$Qwen3.5: assumes the 36T tokens reported for Qwen3. The
    dashed arrows mark larger budgets implied by family-level claims; they are
    not verified bounds.
  \item $^{\S}$Qwen3.5 4B: the score is IRT-estimated.
\end{itemize}

Scores for Qwen3.5 and Nemotron 3 Super are from MathArena, scores for
Nemotron 3 Nano and Nemotron Cascade 2 are from NVIDIA, and the rest are our
own evaluations. Evaluation setups differ across these sources.

\providecommand{\stagegain}[1]{\textcolor{pgtwilight}{#1}}

\providecommand{\stagegain}[1]{\textcolor{pgtwilight}{#1}}

\section{Detailed Results by Context Length}\label{app:context-length-results}

\begin{table}[H]
\centering
\pgcaptionfont
\setlength{\tabcolsep}{4pt}
\renewcommand{\arraystretch}{1.15}

\caption{Benchmark accuracy (\%) across post-training stages at a
64k context length. Gains are measured in percentage points relative
to the immediately preceding checkpoint at the same context length.
\textsuperscript{\(\dagger\)} SFT post-RL scores and gains are offline
estimates from rollouts generated at 128k: responses were retokenized,
truncated to 61,440 tokens, and evaluated with the same scorers.
No new rollouts were generated at 64k. Original token sequences
were not saved, and generation with a different limit may produce
different responses. DPO and RL evaluations also used different
concurrency settings between the 64k and 128k runs.}
\label{tab:post-training-64k}
\begin{tabularx}{\linewidth}{@{}Xr*{4}{rr}@{}}
\toprule
Benchmark &
\multicolumn{1}{c}{\shortstack{Off-policy\\distillation}} &
\multicolumn{2}{c}{\shortstack{DPO\\pre-RL}} &
\multicolumn{2}{c}{\shortstack{First\\RL}} &
\multicolumn{2}{c}{\shortstack{Second\\RL}} &
\multicolumn{2}{c@{}}{\shortstack{SFT\\post-RL\textsuperscript{\(\dagger\)}}} \\
\cmidrule(lr){2-2}\cmidrule(lr){3-4}\cmidrule(lr){5-6}
\cmidrule(lr){7-8}\cmidrule(l){9-10}
& Score & Score & Gain & Score & Gain
& Score & Gain & Score & Gain \\
\midrule[0.9pt]
AIME 2025
& 77.08
& 80.52 & \stagegain{+3.44}
& 85.62 & \stagegain{+5.10}
& 87.71 & \stagegain{+2.09}
& \(\approx 87.40\) & \stagegain{\(\approx -0.31\)} \\

AIME 2026
& 81.67
& 83.54 & \stagegain{+1.87}
& 87.50 & \stagegain{+3.96}
& 87.40 & \stagegain{$-$0.10}
& \(\approx 90.42\) & \stagegain{\(\approx +3.02\)} \\

HMMT Feb.\ 2025
& 70.42
& 76.15 & \stagegain{+5.73}
& 80.73 & \stagegain{+4.58}
& 80.62 & \stagegain{$-$0.11}
& \(\approx 85.94\) & \stagegain{\(\approx +5.32\)} \\

HMMT Feb.\ 2026
& 64.68
& 62.41 & \stagegain{$-$2.27}
& 73.01 & \stagegain{+10.60}
& 72.25 & \stagegain{$-$0.76}
& \(\approx 75.66\) & \stagegain{\(\approx +3.41\)} \\

APEX Shortlist
& 23.27
& 21.81 & \stagegain{$-$1.46}
& 33.44 & \stagegain{+11.63}
& 32.05 & \stagegain{$-$1.39}
& \(\approx 38.16\) & \stagegain{\(\approx +6.11\)} \\

BeyondAIME
& 57.16
& 52.25 & \stagegain{$-$4.91}
& 65.28 & \stagegain{+13.03}
& 65.34 & \stagegain{+0.06}
& \(\approx 67.78\) & \stagegain{\(\approx +2.44\)} \\

ArXivMath May 2026
& 20.08
& 15.39 & \stagegain{$-$4.69}
& 25.08 & \stagegain{+9.69}
& 24.14 & \stagegain{$-$0.94}
& \(\approx 21.09\) & \stagegain{\(\approx -3.05\)} \\
\bottomrule
\end{tabularx}

\par\vspace{12pt}

\caption{Benchmark accuracy (\%) across post-training stages at a
128k context length. Gains are measured in percentage points relative
to the immediately preceding checkpoint at the same context length.
DPO and RL evaluations also used different concurrency settings
between the 64k and 128k runs.}
\label{tab:post-training-128k}
\begin{tabularx}{\linewidth}{@{}Xr*{4}{rr}@{}}
\toprule
Benchmark &
\multicolumn{1}{c}{\shortstack{Off-policy\\distillation}} &
\multicolumn{2}{c}{\shortstack{DPO\\pre-RL}} &
\multicolumn{2}{c}{\shortstack{First\\RL}} &
\multicolumn{2}{c}{\shortstack{Second\\RL}} &
\multicolumn{2}{c@{}}{\shortstack{SFT\\post-RL}} \\
\cmidrule(lr){2-2}\cmidrule(lr){3-4}\cmidrule(lr){5-6}
\cmidrule(lr){7-8}\cmidrule(l){9-10}
& Score & Score & Gain & Score & Gain
& Score & Gain & Score & Gain \\
\midrule[0.9pt]
AIME 2025
& 79.58
& 87.29 & \stagegain{+7.71}
& 85.42 & \stagegain{$-$1.87}
& 87.40 & \stagegain{+1.98}
& 89.17 & \stagegain{+1.77} \\

AIME 2026
& 81.77
& 91.25 & \stagegain{+9.48}
& 88.54 & \stagegain{$-$2.71}
& 86.98 & \stagegain{$-$1.56}
& 91.35 & \stagegain{+4.37} \\

HMMT Feb.\ 2025
& 72.40
& 81.77 & \stagegain{+9.37}
& 81.98 & \stagegain{+0.21}
& 79.48 & \stagegain{$-$2.50}
& 89.69 & \stagegain{+10.21} \\

HMMT Feb.\ 2026
& 66.48
& 74.15 & \stagegain{+7.67}
& 72.16 & \stagegain{$-$1.99}
& 72.16 & \stagegain{0.00}
& 80.78 & \stagegain{+8.62} \\

APEX Shortlist
& 25.33
& 34.51 & \stagegain{+9.18}
& 35.17 & \stagegain{+0.66}
& 30.12 & \stagegain{$-$5.05}
& 43.42 & \stagegain{+13.30} \\

BeyondAIME
& 59.41
& 66.03 & \stagegain{+6.62}
& 66.03 & \stagegain{0.00}
& 65.34 & \stagegain{$-$0.69}
& 72.38 & \stagegain{+7.04} \\

ArXivMath May 2026
& 20.47
& 24.69 & \stagegain{+4.22}
& 26.09 & \stagegain{+1.40}
& 25.86 & \stagegain{$-$0.23}
& 23.52 & \stagegain{$-$2.34} \\
\bottomrule
\end{tabularx}

\end{table}

\begingroup
\newcommand{\scoreSE}[2]{%
  \shortstack[c]{#1\\{\scriptsize\(\pm #2\)}}%
}

\begin{table}[H]
\centering
\pgcaptionfont
\setlength{\tabcolsep}{4pt}
\caption{Standard errors (percentage points) of the pass@1 scores in
Table~\ref{tab:math-evaluation}. For each of the \(R=32\) generation repetitions, we compute accuracy over the full benchmark; the SE is the sample standard deviation of these accuracies divided by \(\sqrt{R}\).
\(\dagger\) denotes externally reported scores, for which no SE is available.}
\label{tab:math-evaluation-sem}
\begin{tabularx}{\linewidth}{@{}>{\raggedright\arraybackslash}Xccccccc@{}}
\toprule
Model & \shortstack[c]{AIME\\2026} &
\shortstack[c]{HMMT\\Feb.\ 2026} & \shortstack[c]{APEX\\Shortlist} &
\shortstack[c]{Beyond\\AIME} & \shortstack[c]{AIME\\2025} &
\shortstack[c]{HMMT\\Feb.\ 2025} &
\shortstack[c]{ArXivMath\\May 2026} \\
\midrule
\rowcolor{pgtwilight!22!white}
\textbf{Limite 1B - Violetto}
  & 0.54 & 0.74 & 0.88 & 1.03 & 0.49 & 0.72 & 0.81 \\
VibeThinker-1.5B
  & 0.95 & 0.81 & 0.56 & 0.37 & 1.08 & 1.20 & 0.62 \\
MiniCPM5-2B
  & 0.65 & 0.95 & 0.99 & 0.53 & 0.75 & 0.92 & 0.56 \\
VibeThinker-3B
  & 0.52 & 0.72 & 1.24 & 0.40 & 0.53 & 0.79 & 0.91 \\
Qwen3.5-4B
  & \(\dagger\) & \(\dagger\) & \(\dagger\) & 1.11 & 0.87 & 1.11 & 0.51 \\
Olmo-3-7B-Think
  & 1.17 & 0.70 & 0.54 & 0.40 & 0.98 & 0.89 & 0.56 \\
Olmo-3-7B-Instruct
  & 1.18 & 0.59 & 0.49 & 0.54 & 1.00 & 0.91 & 0.57 \\
OLMo-Hybrid-Think-SFT-7B
  & 0.86 & 0.71 & 0.57 & 0.46 & 1.07 & 0.91 & 0.46 \\
Qwen3.5-9B
  & 0.61 & 0.74 & 0.79 & 0.46 & 0.76 & 0.72 & 0.73 \\
\midrule
MUSE-Glimmer-30B
  & 0.44 & 0.64 & 1.00 & 0.82 & 0.46 & 0.72 & 0.74 \\
Gemma-4-31B-IT
  & 0.52 & 0.56 & 0.85 & 1.78 & 0.81 & 0.78 & 0.55 \\
\bottomrule
\end{tabularx}
\end{table}

\section{Qualitative examples}
\label{app:qualitative-examples}

\definecolor{chatviolet}{HTML}{E8DDF7}

\newcommand{\limitechatturn}[4]{%
  \par\noindent
  \makebox[\linewidth][#1]{%
    \begingroup
    \setlength{\fboxsep}{7pt}%
    \fcolorbox{black!25}{#2}{%
      \parbox{\dimexpr0.88\linewidth-2\fboxsep-2\fboxrule\relax}{%
        \textbf{#3}\par\smallskip
        #4%
      }%
    }%
    \endgroup
  }%
  \par\smallskip
}

\newcommand{\limiteexample}[4]{%
  \par\medskip
  \noindent\begin{minipage}{\linewidth}
    {\bfseries Example #1: #2}\par\smallskip
    \limitechatturn{l}{black!4}{User:}{#3}
    \limitechatturn{r}{chatviolet}{Limite:}{#4}
  \end{minipage}
  \par\medskip
}

The following selected, independent single-turn interactions illustrate
both Limite's mathematical reasoning and a limitation in instruction
following. They are illustrative examples, not an estimate of how often
either behavior occurs. The original exchanges appear in the release
post.\footnote{\url{https://paradigma.inc/blog/limite-1b-violetto/}}

\subsection{Limitation in instruction following}

\limiteexample{1}{Photosynthesis}{%
What is photosynthesis, in simple terms?%
}{%
A cell divides into two cells, and those cells can each divide
into two\ldots%
}

\subsection{A hard problem, solved}

\limiteexample{2}{A path through a board}{%
Consider a $2025\times2025$ board whose squares are identified with
pairs $(i,j)$, where $i$ and $j$ denote the row and column.

Calvin picks two positive integers $a,b<2025$ and places a pawn at
$(1,1)$. On odd-numbered moves, he moves from $(i,j)$ to either
$(i+a,j)$ or $(i,j+a)$. On even-numbered moves, he moves to either
$(i+b,j)$ or $(i,j+b)$. All coordinates are taken modulo $2025$.

For how many pairs $(a,b)$ can the pawn visit every square exactly
once?%
}{%
Thus a necessary condition is
\[
a=b\quad\text{and}\quad\gcd(a,2025)=1.
\]

\noindent[\ldots]

Write $t=kn+s$ with $0\leq k,s\leq n-1$. Then $R_t=k$ and
\[
y_t\equiv kn+s-k\equiv s-k\pmod n.
\]
Thus the visited vertex is $(k,\;s-k\pmod n)$. As $(k,s)$ runs over
all $n^2$ pairs, $(k,s-k)$ also runs over all $n^2$ pairs, so the
path visits every square exactly once.

\noindent[\ldots]

The admissible pairs are precisely $(a,a)$ with
$1\leq a\leq n-1$ and $\gcd(a,n)=1$. The number of such $a$ is
Euler's totient $\varphi(n)$.
\[
\varphi(2025)
=2025\left(1-\frac13\right)\left(1-\frac15\right)
=1080.
\]
\[
\boxed{1080}
\]%
}

\global\pgsectionrunontrue
\section*{Acknowledgements}
We thank Donato Crisostomi for supervision and project management early in the
project, Alessandro Fazio for support with Flywheel development,
Alessandro Lombardo for operational support, and all three of them, along with
Francesco Prignano, for the many exchanges of ideas.

We thank Tony Silveti-Falls and Minh Nhat Nguyen for feedback on a pre-release draft of this report.

We also thank Manfredi Caracciolo for the aura artwork.

\end{document}